\documentclass{article}

\usepackage{microtype}
\usepackage{graphicx}
\usepackage{subcaption}
\usepackage{booktabs} 

\usepackage{hyperref}
\makeatletter
\g@addto@macro{\UrlBreaks}{\do\-}
\makeatother

\usepackage[accepted]{icml2026}

\usepackage{amsmath}
\usepackage{amssymb}
\usepackage{mathtools}
\usepackage{amsthm}
\usepackage{multirow}

\usepackage{enumitem}

\usepackage[capitalize,noabbrev]{cleveref}

\theoremstyle{plain}

\theoremstyle{definition}

\theoremstyle{remark}

\usepackage[textsize=tiny]{todonotes}

\begin{document}

\twocolumn[
  \icmltitle{Circuit Tracing in Autoregressive Protein Language Models}



  \icmlsetsymbol{equal}{*}

  \begin{icmlauthorlist}
    \icmlauthor{Darin Tsui}{yyy}
    \icmlauthor{William Deinzer}{yyy}
    \icmlauthor{Daniel Saeedi}{yyy}
    \icmlauthor{Amirali Aghazadeh}{yyy}
  \end{icmlauthorlist}

  \icmlaffiliation{yyy}{School of Electrical and Computer Engineering, Georgia Institute of Technology, Atltanta, GA}

  \icmlcorrespondingauthor{Amirali Aghazadeh}{amiralia@gatech.edu}

  \icmlkeywords{Machine Learning, ICML, Cross-layer transcoders, Circuit discovery, Protein language models, ProGen3, Mechanistic interpreability, Biological motifs}

  \vskip 0.3in
]



\printAffiliationsAndNotice{}  

\begin{abstract}
Protein language models (pLMs) can generate novel protein sequences with properties beyond those observed in nature, yet the mechanisms underlying protein generation remain poorly understood. Existing mechanistic interpretability methods based on sparse autoencoders and transcoders primarily focus on protein representation learning models and do not capture the computation required for autoregressive generation. Here, we introduce ProGenMech, a mechanistic interpretability framework for generative protein language models that extends cross-layer transcoders (CLTs) to ProGen3, a sparse Mixture-of-Experts model trained for both causal generation and span infilling. Unlike per-layer approaches, CLTs reconstruct each layer using sparse latent variables from all preceding layers, enabling faithful recovery of inter-layer generative computation. We further develop a zero-shot circuit discovery framework to identify sparse latent circuits responsible for protein generation and fitness prediction. In causal generation and zero-shot fitness estimation tasks, ProGenMech outperforms local transcoder baselines in recovering ProGen3’s probability distribution and functional scoring behavior, while matching the original model’s generative distribution in span infilling tasks. Moreover, the recovered circuits reveal biologically meaningful motifs and functional regions associated with conserved sequence patterns and protein fitness landscapes, establishing a foundation for interpretable and steerable protein generation.

\end{abstract}

\section{Introduction}

\begin{figure*}[t!]
\vspace{-0cm}
\centering
\includegraphics[width=1\textwidth]{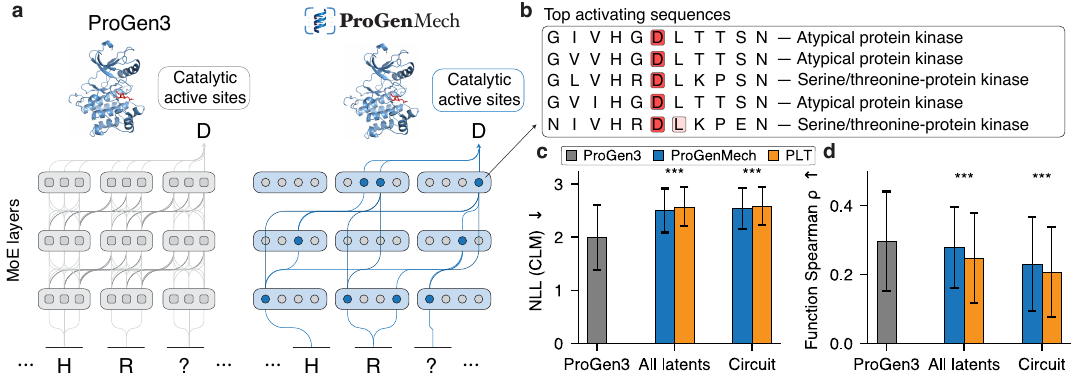}
\vspace{-0.3cm}
\caption{{\bf ProGenMech serves as a replacement model for ProGen3.} {\bf a}, Schematic of the circuit discovery process. ProGenMech identifies sparse circuits of interpretable latents (blue) that trace and approximate ProGen3’s generative computation across causal generation, span infilling, and fitness prediction tasks. {\bf b}, Example of top activating sequences in Swiss-Prot for a specific latent, revealing a conserved catalytic motif associated with protein kinase activity. On {\bf c}, generation and {\bf d}, zero-shot fitness prediction tasks, ProGenMech outperforms PLT baselines in recovering ProGen3’s probability distribution and functional behavior.
} 
\label{fig:circuit_tracing}
\end{figure*}

Protein language models (pLMs) have rapidly transformed computational biology by learning rich statistical representations from large-scale protein sequence corpora. These models now achieve state-of-the-art performance across diverse tasks including structure prediction~\cite{Abramson2024}, fitness estimation~\cite{lin2023evolutionary}, and protein design~\cite{hayes2025simulating}. More recently, generative pLMs have demonstrated the ability to synthesize novel protein sequences with functional and structural properties not observed in nature, raising the possibility of programmable biological design~\cite{bhatnagar2025scaling}. The success of these approaches suggests that pLMs are encoding for the structural and functional motifs governing protein sequences~\cite{rives2021biological, tsui2025shapzero, tsui2024recovering}. Despite these advances, the internal computational mechanisms underlying protein generation remain poorly understood. Modern generative pLMs contain billions of learned parameters distributed across deep transformer architectures, yet little is known about how these models internally represent biological function, compose structural constraints, or coordinate sequence generation across layers. As a result, protein generation remains largely a black-box process, limiting our ability to reliably steer, diagnose, or biologically interpret model behavior.

Recent work in mechanistic interpretability has sought to address this problem by constructing sparse replacement models that expose the internal computation of transformers through interpretable latent variables. Sparse autoencoders (SAEs)~\cite{templeton2024scaling, gao2025scaling} have emerged as a dominant approach for decomposing pLM activations into interpretable features~\cite{adams2025mechanistic, simon2025interplm, walton2025golf, gujral2025sparse, nainani2025mechanistic, parsan2025towards, tsui2025lownsae, corominas2025sparse, garcia2025interpreting}. However, SAEs provide a factorization of representations but do not capture computations across layers. To model these computations, transcoders approximate each transformer layer through a sparse latent bottleneck, enabling the construction of sparse replacement models~\cite{dunefsky2024transcoders}. Per-layer transcoders (PLTs) reconstruct each layer independently; however, they fail to capture the accumulation of context and computation across depth~\cite{ameisen2025circuit}. 

Cross-layer transcoders (CLTs) address this limitation by reconstructing each layer as a function of sparse latent variables from all preceding layers, enabling a globally connected replacement model that preserves inter-layer computation~\cite{ameisen2025circuit}. In pLMs, the ProtoMech framework recently demonstrated that CLTs can recover biologically meaningful circuits in ESM2, exposing latent motifs associated with protein families and function prediction while preserving a substantial fraction of model behavior~\cite{tsui2026protomech}. However, ESM2 is a masked representation model rather than a generative model. Consequently, ProtoMech cannot expose the circuits that govern protein generation. 

Here, we introduce ProGenMech, a CLT-based mechanistic interpretability framework for generative protein language models, by extending cross-layer transcoders to ProGen3~\cite{bhatnagar2025scaling}, a sparse Mixture-of-Experts generative pLM\footnote{Our code can be found at \url{https://github.com/amirgroup-codes/ProGenMech}, and our visualizer is available at \url{https://protmech.github.io/}.}. We develop architectural and training modifications that enable CLTs to faithfully approximate ProGen3’s generative computation across both causal generation and span infilling regimes, and introduce a zero-shot circuit discovery framework to identify sparse latent circuits responsible for protein generation and fitness prediction. Our results demonstrate that ProGenMech outperforms PLT baselines in recovering generative behavior while revealing biologically meaningful circuits associated with conserved motifs, functional regions, and fitness landscapes. This work makes four primary contributions:
\begin{enumerate}[leftmargin=5mm]
    \item We introduce ProGenMech, a cross-layer transcoder framework for mechanistic interpretability in autoregressive protein language models.
    
    \item We develop architectural and training adaptations enabling CLTs to faithfully approximate sparse Mixture-of-Experts generative computation across both causal generation and span infilling objectives.
    
    \item We introduce a zero-shot circuit discovery framework that identifies sparse latent circuits governing protein generation and fitness prediction.
    
    \item We demonstrate that the recovered circuits expose biologically meaningful motifs and functional regions while outperforming per-layer transcoders in recovering generative behavior.
\end{enumerate}

\section{Cross-layer Transcoders (CLTs)}

\begin{figure*}[t!]
\vspace{-0cm}
\centering
\includegraphics[width=1\textwidth]{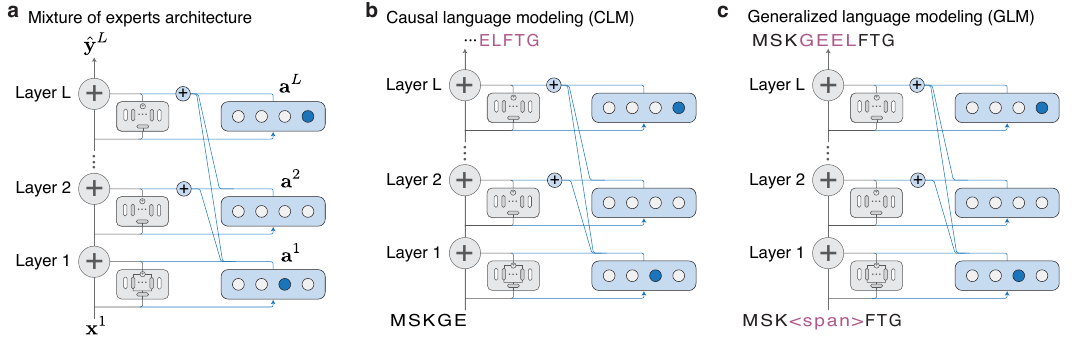}
\vspace{-0.3cm}
\caption{{\bf Adapting CLTs for ProGen3.} {\bf a}, ProGenMech reconstructs the cumulative computation of ProGen3's sparse mixture of experts architecture at every layer. {\bf b}, ProGenMech is trained on both causal language modeling (CLM), where sequences are inputted for left-to-right autoregressive generation, and {\bf c}, generalized language modeling (GLM) tasks, where spans of sequences are infilled.}
\label{fig:overview}
\end{figure*}

We begin by defining the standard CLT architecture. CLTs are extensions of transcoders, which aim to approximate the input–output relationship of each MLP layer as a function of sparse latent variables at all preceding layers (Fig.~\ref{fig:overview}a). Specifically, for a model with $L$ layers and a hidden dimension $d_{\text{model}}$, we denote $\mathbf{x}^{\ell} \in \mathbb{R}^{d_{\text{model}}}$ as the residual stream activation at the $\ell^{\text{th}}$ layer, where $\ell \in {1,\dots,L}$. $\mathbf{x}^{\ell}$ then typically gets passed into an MLP, for which we denote the output as $\mathbf{y}^{\ell} \in \mathbb{R}^{d_{\text{model}}}$.

The objective of a CLT is to reconstruct $\mathbf{y}^{\ell}$ as a function of sparse latent variables derived from all preceding layers $1,\dots,\ell$. To do this, we define an encoder for each layer that maps the residual stream input $\mathbf{x}^{\ell}$ to a sparse latent representation $\mathbf{a}^{\ell} \in \mathbb{R}^{d_{\text{latent}}}$ via:
\begin{equation}
    \mathbf{a}^{\ell} = \text{TopK}(\mathbf{W}_{\text{enc}}^{\ell}(\mathbf{x}^{\ell} - \mathbf{b}_{\text{pre}}^{\ell}) + \mathbf{b}^{\ell}_{\text{enc}}),
    \label{eq:encoder}
\end{equation}
where $\mathbf{W}_{\text{enc}}^{\ell} \in \mathbb{R}^{d_{\text{latent}} \times d_{\text{model}}}$ is the encoder weight matrix, $\mathbf{b}_{\text{pre}}^{\ell} \in \mathbb{R}^{d_{\text{model}}}$ is the corresponding bias term at layer $\ell$, and $\mathbf{b}_{\text{enc}}^{\ell} \in \mathbb{R}^{d_{\text{model}}}$ is the encoder bias at layer $\ell$. To promote sparsity in the latent space, the $\mathrm{TopK}$ operator~\cite{makhzani2013ksparse} retains only the $k$ largest-magnitude latent activations and sets all others to zero. 

To reconstruct $\mathbf{y}^{\ell}$, denoted as $\hat{\mathbf{y}}^{\ell} \in \mathbb{R}^{d_{\text{model}}}$, CLTs employ decoder matrices that map latent representations from preceding layers to layer $\ell$ via:
\vspace{-1mm}
\begin{equation}
    \hat{\mathbf{y}}^{\ell} = \sum_{\ell'=1}^{\ell} \mathbf{W}_{\text{dec}}^{\ell^{'} \rightarrow \ell} \mathbf{a}^{\ell^{'}} + \mathbf{b}_{\text{pre}}^{\ell},
    \label{eq:decoder}
\end{equation}
where $\mathbf{W}_{\text{dec}}^{\ell' \rightarrow \ell}$ is the decoder weight matrix mapping latent features from layer $\ell'$ to layer $\ell$. By requiring each reconstruction to depend on previous layers, CLTs can capture cumulative context and expose the inter-layer pathways that lead to $\mathbf{y}^{\ell}$.

To train the CLT, we minimize the mean-squared error between $\mathbf{y}^{\ell}$ and its reconstruction $\hat{\mathbf{y}}^{\ell}$ across all layers,
$\mathcal{L}_{\text{MSE}} = \sum_{\ell=1}^{L} \| \mathbf{y}^{\ell} - \hat{\mathbf{y}}^{\ell} \|_2^2$. To reduce the number of inactive (“dead”) latents during training, we incorporate an auxiliary loss inspired by~\cite{gao2025scaling, tsui2026protomech}. We denote $\mathbf{e}^{\ell} = \mathbf{y}^{\ell} - \hat{\mathbf{y}}^{\ell}$ as the reconstruction residual at layer $\ell$. We then define the auxiliary loss: $    \mathcal{L}_{\text{aux}} = \sum_{\ell=1}^{L} \| \mathbf{e}^{\ell} - \hat{\mathbf{e}}^{\ell} \|_2^2.$ $\hat{\mathbf{e}}^{\ell}$ is obtained by decoding the top-$k_{\text{aux}}$ latent activations in $\mathbf{a}^{\ell}$ using the decoder matrix $\mathbf{W}_{\text{dec}}^{\ell \rightarrow \ell}$, where $k_{\text{aux}}$ is a hyperparameter. The full CLT training objective is defined as: $\mathcal{L}_{\text{CLT}} = \mathcal{L}_{\text{MSE}} + \alpha \mathcal{L}_{\text{aux}},$
where $\alpha$ controls the weight of the auxiliary loss. This joint objective enforces faithful reconstruction while encouraging broad usage of the latent space.

We train CLTs on the ProGen3-112M model, for which $L = 10$ and $d_{\text{model}} = 384$. Following~\cite{tsui2026protomech}, we train on 5 million protein sequences of length up to 1022 amino acids, randomly sampled from UniRef50~\cite{suzek2007uniref}. After training, the CLT functions as a replacement model for ProGen3. At each layer, the CLT substitutes the original MLP block with a sparse, interpretable latent representation that approximates the original computation (Fig.~\ref{fig:circuit_tracing}a). We refer to Appendix~\ref{app:model} for additional hyperparameter considerations. To enable controlled comparison, we also train a PLT using the same hyperparameters and sparsity constraint $k$ as the CLT (see Appendix~\ref{app:plt}).

\subsection{Building CLTs for ProGen3}

To construct a faithful replacement model for ProGen3, the CLT architecture must be adapted to account for the model’s unique generative capabilities. Unlike standard transformers, ProGen3 uses a sparse Mixture of Experts (MoE) framework and a dual-task training objective, which complicates the direct application of CLTs. In this section, we detail the specific architectural and training modifications we make to adapt the CLT framework for ProGen3.

\textbf{Sparse Mixture of Experts (MoE) Integration.} Unlike traditional transformers, which use MLP layers, ProGen3 utilizes a sparse MoE architecture where residual stream activations at each layer are dynamically routed to a subset of expert sub-networks. To account for this, we set $\mathbf{x}^{\ell}$ as the input to the MoE layer at layer $\ell$ and $\mathbf{y}^{\ell}$ as the final aggregated output of the expert sub-networks (Fig.~\ref{fig:overview}a). By approximating the entire MoE operation as a single functional mapping, the CLT discovers a unified, sparse latent space that captures the overall computation of the experts.

\textbf{Generative and Infilling Training Objectives.} ProGen3 is trained using a hybrid objective that encompasses both causal and generalized language modeling. In causal language modeling (CLM), ProGen3 does left-to-right autoregressive generation, conditioning on the previous amino acids in the sequence (Fig.~\ref{fig:overview}b). In generalized language modeling (GLM), ProGen3 does span infilling, where a portion of the sequence is masked out, requiring ProGen3 to infill the masked out amino acids (Fig.~\ref{fig:overview}c). To account for both modes, we follow the training procedure of ProGen3 and incorporate a 2:1 training split between CLM and GLM sequences, meaning that in our training set, 1/3rd of the sequences seen by the model are partially masked out~\cite{bhatnagar2025scaling}. To determine the span length masked, we randomly sample from a mixture of five Gaussian distributions with equal probability, $\mathcal{N}(10, 5), \mathcal{N}(30, 10), \mathcal{N}(70, 20), \mathcal{N}(200, 50),$ and $\mathcal{N}(400, 100)$. The maximum fraction of the sequence masked out is determined by sampling from the set $\{0.15, 0.25, 0.5, 0.8\}$ with probabilities $\{0.28, 0.3, 0.28, 0.14\}$, respectively. This exposure to diverse masking states ensures that the CLT is able to represent ProGen3's full generative capabilities. 

\section{Circuit Discovery}

Leveraging the interpretable latent space learned by the CLT, we aim to isolate the minimal subset of latent variables that governs computation in ProGen3. Following~\cite{dunefsky2024transcoders}, we define a \emph{circuit} as a set of latent variables whose activity is responsible for a specific model behavior. While the original ProtoMech framework utilized supervised probes to identify circuits for classification and regression in ESM2~\cite{tsui2026protomech}, this approach is fundamentally limited when applied to autoregressive pLMs. In this work, we pivot to a zero-shot circuit discovery paradigm, where the target computation is the model’s internal probability distribution for generation and scoring.

\textbf{Replacement Model.} To identify such circuits, we first compute the original model’s logits from $\mathbf{y}^L$ to establish a baseline for generative and scoring performance. We then seek the minimal subset of latent variables capable of producing logits derived from the reconstructed final-layer output $\hat{\mathbf{y}}^{L}$ that recovers this performance. 

To compute $\hat{\mathbf{y}}^{L}$, we adopt a replacement model similar to~\cite{ameisen2025circuit}, where during
the forward pass, the original MoE outputs are substituted by their CLT circuit reconstructions, while the attention head outputs are held fixed to those of the ground-truth ProGen3 model. This formulation isolates the contribution of
the MoE layers while avoiding error accumulation associated with reconstruction attention representations. 

Similar to findings from~\cite{ameisen2025circuit, tsui2026protomech}, we observe that a fully recursive replacement, where both MoE and attention computations are derived from CLT reconstructions at each layer, leads to substantial performance degradation due to compounding reconstruction errors from replacing the attention mechanism. As such, we benchmark our CLT and PLT using this fixed attention-head strategy in the main text. We additionally benchmark against other replacement models tested on in ProtoMech, detailed in Appendix~\ref{app:replacement}.

\subsection{Circuit Discovery Tasks}
\label{sec:tasks}

In this section, we detail the circuit discovery tasks used to benchmark our CLT. We focus on three tasks most representative of ProGen3’s utility: CLM and GLM generation and zero-shot function prediction. 

\textbf{CLM and GLM Generation.} For CLM and GLM generative tasks, the circuit’s objective is to faithfully recover ProGen3's learned probability distribution when generating many sequences. In CLM generation, we provide ProGen3 with the context of 80\% of a real-world biological sequence and autoregressively generate the remaining amino acids until the original sequence length is reached. In GLM generation, we task ProGen3 with infilling a masked span within a real-world sequence. The span length is determined by randomly sampling from the two smallest mixture of Gaussian distributions, $\mathcal{N}(10, 5), \mathcal{N}(30, 10)$, with equal probability.

To construct these tasks, we utilize Swiss-Prot sequences~\cite{boeckmann2003swissprot} clustered at 30\% sequence identity to ensure structural diversity. From the Swiss-Prot 30\% identity dataset, we randomly sample 1000 sequences for circuit discovery. Following a similar procedure to ProGen3's experimental setup, to mitigate the prevalence of low-quality, repetitive artifacts common in language model generations, we ensure that each generated sequence contains less than 25\% low-complexity regions~\cite{bhatnagar2025scaling}, which we define as regions consisting of tandem repeats where the total consecutive repetitive span is at least six residues in length.

During generation, we employ top-$p$ sampling with $p=0.95$ and $T=0.5$. We select a low temperature $T$ to highlight the distributional shifts between ProGen3 and CLT and PLT circuits. To prevent temporal error accumulation and compounding reconstruction drift in circuits, at each generative step, the replacement model is provided with the ground-truth ProGen3 activations from the preceding step.

We note that in the NLP literature, circuits are typically benchmarked on their ability to reconstruct a single next token, often measured by top-1 accuracy or KL divergence~\cite{ameisen2025circuit}. However, for pLMs, the utility of the model lies in its ability to generate functionally plausible sequences rather than emulating a single ground-truth token at a specific position. Consequently, we evaluate the performance of our circuits by scoring the generated portions of the sequence using the original ProGen3 model's negative log-likelihood (NLL) as a proxy for function.

\textbf{Zero-shot Function Prediction.} Beyond generation, we investigate the circuits driving ProGen3’s ability to predict zero-shot fitness. To this, we select eight Deep Mutational Scanning (DMS) assays from ProteinGym~\cite{notin2023proteingym}, spanning a diverse set of biological functions (see Table~\ref{function-circuit-table} for the full list of assays). Following the zero-shot scoring protocol in ProGen3, we compute the zero-shot fitness score as the average log-likelihood between scoring the sequence in the forward and reverse directions. Due to computational constraints, from each DMS assay, we randomly sample 1000 sequences and compute the Spearman correlation between the log-likelihoods and experimental fitness values in ProGen3 and the CLT and PLT circuits reconstructions. 

\subsection{Circuit Discovery Algorithm} Following a similar procedure to~\cite{nainani2025mechanistic, dunefsky2024transcoders}, we employ an iterative greedy search procedure based on gradient-based attribution to recover the minimal subset of latent variables required for each task. For each latent, we compute an attribution score by measuring its contribution toward minimizing the mean KL divergence between the original and reconstructed logits. In CLM and GLM generation, we measure the mean KL divergence in only the logits of the generated tokens. In zero-shot function prediction, we measure the mean KL divergence of the entire logit matrix. We then rank the latents by attribution magnitude and incrementally add latents in small batches to the candidate circuit. This procedure continues until the resulting circuit's mean KL divergence is $\le 1.2\times$ the divergence of the full CLT baseline in CLM and GLM generation. In zero-shot function prediction, this procedure continues until the circuit recovers at least 70\% of the original ProGen3 Spearman correlation or reaches the performance of the replacement model when all latents are included. Additional details of the circuit discovery procedure are provided in Appendix~\ref{app:experimental-setup}.

\subsection{Visualizing Circuits}

To qualitatively analyze the computational pathways discovered, we adapt the ProtoMech visualizer for ProGen3\footnote{We have integrated ProGen3 into the ProtoMech visualizer at \url{https://protmech.github.io/}}. We represent each circuit as a graph where nodes correspond to interpretable latents and edges represent the virtual influence between them. For a given input sequence, we isolate the graph by selecting the top-five highest activating nodes per layer, as determined by their task-specific attribution scores. When analyzing fitness landscapes, we compare the wildtype sequence against high- and low-fitness variants by additionally including the top five nodes with the largest activation shift between the wildtype and the mutant.

From here, we seek to identify possible biological overlap learned by each node. We identify the specific amino acids that drive latent activation in the input sequence and cross-reference them with the top-10 maximally activating sequences from the Swiss-Prot database to detect conserved motifs. These activations are subsequently projected onto the protein structure to visualize their spatial distribution and proximity to known functional sites. Finally, to trace the computational logic governing the model's output, we define virtual edges between latents in subsequent layers following~\cite{ameisen2025circuit}. We calculate the weight of an edge $A_{s \rightarrow t}$ connecting a source node $s$ to a target node $t$ as the product of the source node's activation and the gradient of the target node’s pre-activation with respect to the source. Formally, we define $A_{s \rightarrow t}$ as:
$$\begin{aligned}
A_{s \rightarrow t} &= a_s w_{s \rightarrow t} \\
&= a_s \sum_{\ell_s \leq \ell \leq \ell_t}(W_{\text{dec},s}^{\ell_s \rightarrow \ell})^T J^{\blacktriangledown}_{t_s, \ell_s \to t_t, \ell_t} v_{in,t},
\end{aligned}$$
where $J^{\blacktriangledown}_{t_s, \ell_s \to t_t, \ell_t}$ represents the Jacobian between token $s$ at layer $\ell_s$ and token $t$ at layer $\ell_t,$ with frozen attention patterns and layernorm denominators. When the target node is a latent at a layer, $v_{in,t} = W_{\text{enc},t}^{\ell_t}$. When the target node is an output logit, we instead define $v_{in,t}$ as the gradient of the target logit minus the mean logit across the vocabulary. Intuitively, $w_{s \rightarrow t}$ acts as a linear weight that measures the gradient between the source activation $a_s$ and the target pre-activation $z_t^{\ell_t},$ where $\mathbf{z}^{\ell_t} \coloneqq \mathbf{W}_{\text{enc}}^{\ell_t}(\mathbf{x}^{\ell_t} - \mathbf{b}_{\text{pre}}^{\ell_t}) + \mathbf{b}_{\text{enc}}^{\ell_t}.$ For more details about the visualizer, we refer to Appendix~\ref{app:viz_tool}.

\section{Results}

\begin{figure*}[t!]
\vspace{-0cm}
\centering
\includegraphics[width=0.85\textwidth]{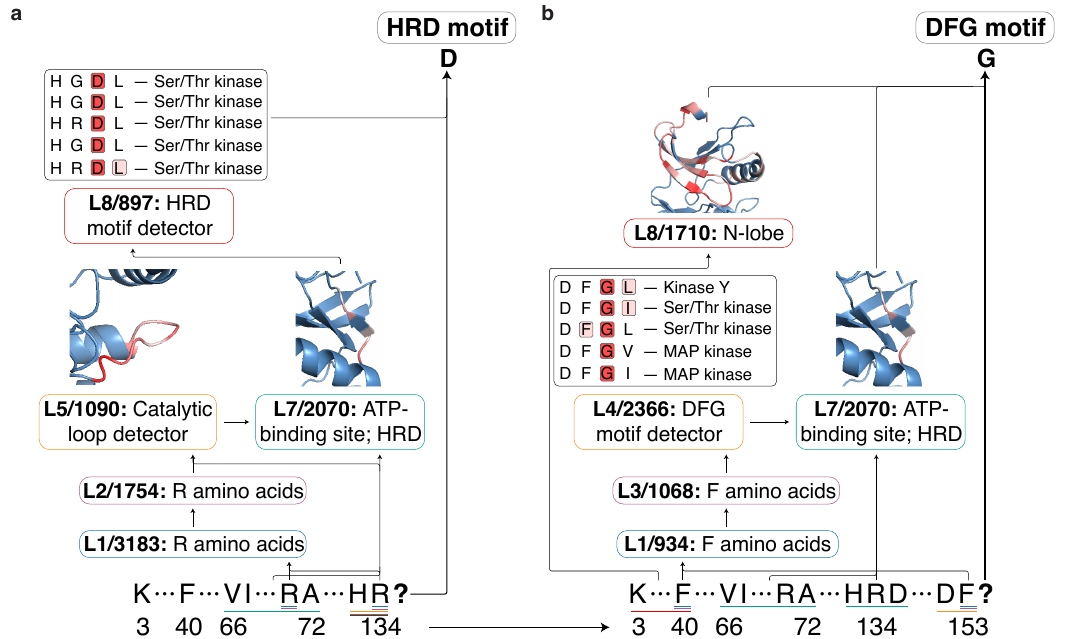}
\vspace{-0cm}
\caption{{\bf Examples of CLM circuits discovered using ProGenMech.} We use the ProGenMech visualization tool to examine generating a kinase protein at the {\bf a}, HRD motif and at the  {\bf b}, DFG motif. We find interpretable features relating to the function of both motifs.
} 
\label{fig:circuits}
\end{figure*}

In this section, we evaluate the performance of ProGenMech on three tasks: \emph{(i)} generative circuit discovery in CLM and GLM settings, \emph{(ii)} zero-shot fitness circuit discovery, and \emph{(iii)} uncovering overlapping biological motifs in circuits. For additional experimental results, we refer to Appendix~\ref{app:expt_results}.

\subsection{ProGenMech Approximates ProGen3's Generation}
\label{sec:generation_results}

We first evaluate ProGenMech on its ability to generate sequences in the CLM setting (Fig.~\ref{fig:circuit_tracing}c). Using the full set of latents, ProGenMech achieves an average NLL of $2.50 \pm 0.42$, recovering approximately 60\% of the original model's likelihood ($2.00 \pm 0.62$), defined as $e^{\text{NLL}_{PG3} - \text{NLL}_{rep}} \times 100$, where $\text{NLL}_{PG3}$ is the NLL of ProGen3 and $\text{NLL}_{rep}$ is the NLL of the replacement model. This significantly outperforms the PLT baseline, which achieves an average NLL of $2.57 \pm 0.36$, suggesting that representation learned by the ProGenMech provides a better approximation of ProGen3's learned distribution. 

This performance advantage continues in circuit discovery. In ProGenMech, the circuits identified achieve an average NLL of $2.54 \pm 0.39$ (58\% likelihood recovery), compared to $2.59 \pm 0.36$ in the PLT case. Notably, ProGenMech achieves this while using $719 \pm 339$ latents (less than 2\% of the total latent space), demonstrating its ability to compress ProGen3's computation. 

We then move to evaluate ProGenMech in the GLM setting. Interestingly, we observe that the NLL distributions for both the ProGenMech and PLT models closely match the original model's baseline. With the full latent set, the ProGenMech and PLT achieve an average NLL of $2.89 \pm 0.46$ and $2.87 \pm 0.43$, respectively, compared to the original model's average NLL of $2.91 \pm 0.44$. In circuit discovery, ProGenMech and PLT circuits achieve an average NLL of $2.89 \pm 0.45$ and $2.88 \pm 0.41$. 

We attribute this phenomenon to the fact that ProGen3 innately struggles to generate functional sequences in GLM generation in the 112M model. ProGen3 exhibits significantly higher average NLL in the GLM setting ($2.91$) compared to the CLM setting ($2.00$), suggesting that infilling remains a challenging domain for models of this scale. As such, in our qualitative analysis, we particularly focus on identifying overlapping biological motifs in the CLM setting, rather than the GLM setting. We anticipate that as the ProGenMech is scaled to larger ProGen3 variants (e.g., the 219M or 339M parameters), the ProGenMech will again yield the performance gains compared to the PLT observed in the CLM setting.

\subsection{ProGenMech Approximates Zero-shot Fitness Prediction}

We observe similar trends when applying ProGenMech to zero-shot fitness tasks (Fig.~\ref{fig:circuit_tracing}d). Using the full set of latents, ProGenMech achieves an average Spearman correlation of $0.28 \pm 0.12$, recovering approximately 95\% of the original model's performance ($0.29 \pm 0.15$). ProGenMech again outperforms the PLT baseline, which achieved an average Spearman correlation of $0.25 \pm 0.12$. 

This performance advantage is reflected similarly in circuit discovery. Here, ProGenMech identified circuits achieve an average Spearman correlation of $0.23 \pm 0.13$ (80\% performance recovery), compared to $0.22 \pm 0.12$ in the PLT baseline. ProGenMech achieves this while using $256 \pm 334$ latents, making up 0.6\% of the latent space. 

\subsection{ProGenMech Reveals Known Biological Motifs}

\begin{figure*}[t!]
\vspace{-0cm}
\centering
\includegraphics[width=0.85\textwidth]{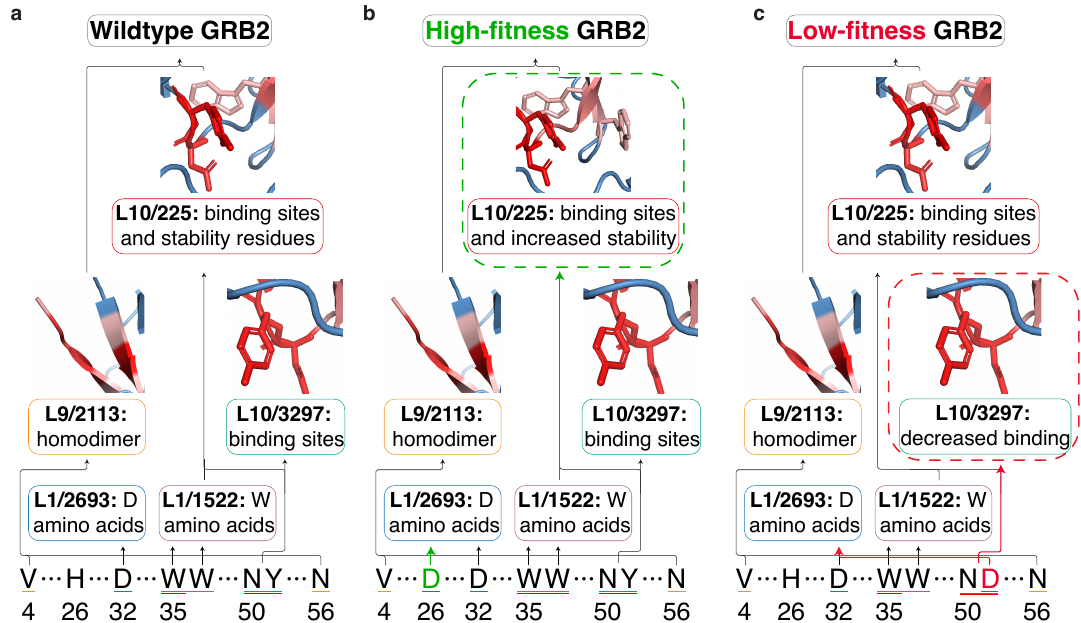}
\vspace{-0cm}
\caption{{\bf Examples of zero-shot fitness circuits discovered using ProGenMech.} {\bf a}, We feed in the wildtype sequence of GRB2 into ProGenMech and identify latents related to binding affinity and stability. {\bf b}, A high-fitness variant of GRB2 increases the activation of a latent associated with stability. {\bf c}, Conversely, a low-fitness variant of GRB2 decreases the activation of a latent associated with binding. 
} 
\label{fig:function_circuits}
\end{figure*}

Given the performance of ProGenMech, we now aim to trace the computational pathways of ProGen3 to uncover known biological motifs. In this section, we perform a rigorous qualitative analysis on our circuits, focusing on the CLM generation of kinase at important motifs and zero-shot scoring the GRB2 fitness landscape. Our analysis reveals that ProGenMech successfully reveals computational pathways in ProGen3 that overlap with known biological motifs.

\textbf{CLM Generation of Kinase.} Fig.~\ref{fig:circuits} depicts circuits discovered governing CLM generation of the kinase protein (UniProt ID P83104), which catalyzes the chemical reactions that regulate cell signaling and behavior~\cite{taylor2011kinase}. We specifically focus on two regions of kinase which attribute to its function: the HRD motif (located at positions 133-135), which is a part of the highly conserved catalytic loop responsible for binding substrate peptides for phosphorylation, the key mechanism for kinase signaling~\cite{modi2019kinase}, and the DFG motif (located at positions 152-154), which forms part of the ATP-binding site and controls whether kinase is in its active or inactive state~\cite{Vijayan2014}.

Fig.~\ref{fig:circuits}a analyzes the CLM generation of the HRD motif. Consistent with findings in~\cite{tsui2026protomech}, we observe that earlier layers tend to recognize specific amino acids that lead to more complex motifs in later layers. Specifically, layer 1 at latent 3183 (L1/3183) and layer 2 at latent 1754 (L2/1754) activate on arginine (R) amino acids, which serve as essential building blocks for catalytic activity~\cite{kornev2006kinase}. This feeds directly into layer 5 (L5/1090), which identifies the general conserved catalytic loop. From here, this information is fed into layer 7 (L7/2070), which narrows the context to the HRD motif within the catalytic loop and highlights parts of the ATP-binding site with which the HRD motif interacts. Lastly, we observe that layer 8 (L8/897), which also has the largest attribution score in the circuit, is primarily responsible for detecting aspartic acid (D) in the HRD motif. Taken as a whole, this circuit suggests that ProGen3 detects basic biochemical patterns, feeds that information into middle layers, which process more complex functional motifs, and sends it to later layers, which finalize the next-token prediction.

Fig.~\ref{fig:circuits}b analyzes the CLM generation of the DFG motif, which appears after ProGen3 sees the HRD motif. Similar to the HRD circuit, layer 1 (L1/934) and layer 3 (L3/1068) activate on phenylalanine (F) amino acids, which are present in the DFG motif. This then feeds into layer 4 (L4/2366), which is responsible for detecting glycine (G) in the DFG motif. L4/2366 also has the largest attribution score in the circuit. This information is fed into layer 7 (L7/2070), which appears in both the HRD and DFG motif circuits. Separately, layer 8 (L8/1710) also detects the N-terminal lobe of kinase, which helps anchor ATP during catalysis~\cite{modi2019kinase}. Altogether, this suggests that ProGen3 has already made up its mind to generate G by layer 4, but uses the context in earlier and later layers to reinforce its decision.

\textbf{Zero-shot Scoring of GRB2.} Fig.~\ref{fig:function_circuits} depicts circuits from the C-terminal SH3 domain of the human growth factor receptor-bound protein 2 (GRB2) fitness landscape~\cite{faure2022mapping}. GRB2 interacts with other proteins in the human body, and is responsible for cellular signalling~\cite{TARI2001}. We first begin by inputting the wildtype sequence into ProGenMech (Fig.~\ref{fig:function_circuits}a). Similar to the CLM circuits, we find that layer 1 primarily detects basic biochemical patterns, like D amino acids (L1/2693) and tryptophan (W) amino acids (L1/1522). We observe that most of the interpretable latents in GRB2 are located in the later layers of ProGen3. For instance, layer 9 (L9/2113) activates on parts of the homodimer interface, which is responsible for binding to other GRB2 proteins. Layer 10 contains the most latents that can be attributed to the function of GRB2. Here, L10/3297 activates on the binding sites of GRB2 that interact with GRB2-associated-binding protein 2 (GAB2), and L10/225 activates on those same binding sites as well as residues stabilizing the hydrophobic core~\cite{faure2022mapping}.

We then leveraged the GRB2 fitness landscape to investigate the latents driving the prediction of ProGen3 toward high-fitness mutations. We analyze the known high-fitness variant, H26D (histidine to aspartic acid at position 26) (Fig.~\ref{fig:function_circuits}b). As expected, L1/2693 activates on H26D, continuing its role as a D-amino acid detector. Additionally, we observe that the attribution score of L10/225 toward this high-fitness prediction increases by more than 58\%, suggesting that H26D, despite not being a hydrophobic acid, interacts with the hydrophobic core of GRB2, increasing its stability, which is consistent with experimental data.

Lastly, we analyze the known low-fitness variant, Y51D (tyrosine to aspartic acid at position 51) and observe its impact to the circuit (Fig.~\ref{fig:function_circuits}c). Similarly to the high-fitness case, L1/2693 activates on this new D amino acid. Additionally, we observe that L10/3297, which was previously activating on binding sites, has an attribution score that has been reduced by about 57\%. Since position Y51 is a known binding site to GAB2, this matches our experimental data that a decrease in fitness can be attributed to a decrease in binding capability. These examples underscore ProGenMech's ability to provide interpretable, biologically plausible explanations toward high- and low-fitness mutations.

\section{Discussion}

\textbf{Adapting CLTs for MoE Architectures.} A central challenge in interpreting ProGen3 lies in its reliance on a sparse MoE architecture, where computation at each layer is routed to expert sub-networks. In this work, we addressed this complexity by defining the CLT input $\mathbf{x}^{\ell}$ as the pre-MoE residual stream and $\mathbf{y}^{\ell}$ as the final aggregated output of the expert ensemble. While this treatment of the MoE block enabled the discovery of generative and zero-shot fitness circuits and allowed us to identify overlapping biological motifs, it fundamentally abstracts away the internal routing logic that is present in ProGen3. A more principled approach to constructing MoE replacement models may involve the construction of expert-specific latent spaces rather than a single unified latent space. Another exciting future direction may be adapting the crosscoder architecture~\cite{lindsey2024sparsecrosscoders}, a variant of sparse autoencoders designed to jointly model multiple activation spaces that has recently been explored for interpreting MoE models~\cite{Chaudhari2026}, to the CLT framework. By extending the CLT architecture to account for expert-subnetwork activation spaces, future iterations of ProGenMech would allow the replacement model to more faithfully recover the computational mechanisms being routed to different subnetworks.

\textbf{Scaling ProGenMech.} A significant consideration when applying ProGenMech to larger architectures is the parameter overhead associated with training cross-layer transcoders. Due to the CLT's reliance on cross-layer connections, the number of decoder matrices required scales with $\mathcal{O}(L^2)$, whereas the encoder count remains linear, $\mathcal{O}(L)$. In our current configuration, the CLT comprises approximately 115M parameters. However, our findings in the GLM task underscore the necessity of scaling ProGenMech to larger ProGen3 models in order to interpret every facet of the model's capabilities and get closer to an interpretable understanding of state-of-the-art protein sequence generation.

\textbf{Limitations.} One application of developing replacement models is the potential for steering, where latents within a discovered circuit are intervened on to bias the model toward specific behaviors. We evaluated the steerability of our zero-shot fitness circuits by applying an intervention strategy similar to~\cite{ameisen2025circuit}, seeking to generate sequences with improved functional attributes using NLL as a proxy for fitness. However, we observed that the NLL of the steered sequences remained distributionally indistinguishable from those generated naturally by the base ProGen3 model (Table~\ref{tab:ProGen_steering_results}). Upon closer examination, we found that the generated portions of sequences, in both the baseline and steered cases, consistently failed our low-complexity region filter described in Section~\ref{sec:tasks} (Fig.~\ref{fig:sequence_snippets}). This suggests a fundamental discrepancy between the generative and scoring components of the 112M-parameter ProGen3 model. While model likelihoods correlate with fitness in scoring mode, the model lacks the capability to generate sequences within that same fitness distribution. Future work comparing the circuits responsible for scoring versus generation may pinpoint where this functional information is lost. We anticipate that scaling ProGenMech to larger ProGen3 variants will reconcile this discrepancy, potentially enabling steering circuits to design high-functioning sequences.

Another limitation is in the interpretation of the discovered circuits, which requires manually parsing the circuit and cross-referencing the information with biological annotations. As our ability to annotate circuits is bounded by existing biological knowledge, ProGenMech may identify circuits governing mechanisms that are not yet formally characterized. We focused our qualitative analysis on kinase and GRB2, where there exists an abundance of annotations and biological studies done, which allows us to validate the identified biological motifs. Developing automated annotation pipelines remains a critical next step to accelerate the discovery of novel biological insights.

\section*{Impact Statement}

This paper presents work whose goal is to advance the field of Machine Learning, specifically toward interpreting protein language models. The potential societal impacts of our work are overwhelmingly positive. Potential applications of this work include discovering biological mechanisms, enhancing protein design, and improving the performance of downstream tasks.

\section*{Acknowledgment} This research was supported by the National Science Foundation (NSF) Graduate Research Fellowship Program (GRFP), the Parker H. Petit Institute for Bioengineering and Biosciences (IBB) interdisciplinary seed grant, the Exponential Electronics seed grant of the Institute for Matter and Systems (IMS) at Georgia Tech, Microsoft via GT Cloud Hub, Georgia Tech Research Corporation, and Georgia Institute of Technology start-up funds.


\bibliography{reference}
\bibliographystyle{icml2026}

\newpage
\appendix
\onecolumn

\section{Model Architecture and Hyperparameters}
\label{app:model}

We train our models on 5 million random sequences from UniRef50 with up to 1022 residues. For ProGen3-112M, where $L=10$ and $d_{\text{model}}=384$, we collect activations from immediately before and after the sparse MoE blocks. When mapping the residual stream activations into each layer's latent space, we first apply the TopK function. We set $k=64$ for each layer in the CLT and PLT, which we find to be the minimum number of latents active in order to obtain decent reconstruction, resulting in a total sparsity of $64\times10=640$ latents for a single amino acid. Additionally, we set $d_{\text{latent}}=4608$, which is a $12\times$ expansion factor from the hidden dimension of ProGen3-112M. We also set $k_{\text{aux}}= \lfloor d_{\text{model}}/2 \rfloor$ and $\alpha=1/32$ based on ~\cite{gao2025scaling}. We note that the auxk loss serves to avoid dead latents by forcing them to activate, and we activate them using the decoder matrix $\mathbf{W}_{\text{dec}}^{\ell \rightarrow \ell}$ for computational efficiency.

We use a batch size of 16 and learning rate of $2\times10^{-4}$ with the AdamW optimizer to train our models. We also utilize a gradient clipping value of $1$ and weight decay of $1\times10^{-5}$ to help stabilize training. Additionally, we normalize the mean-squared error of each layer by dividing by the variance of the ground-truth residual stream activations in each batch. With these hyperparameters, we obtain an average normalized mean-squared error of $0.25$ on a held-out validation set during training.

\section{Per-layer Transcoders}
\label{app:plt}

We implement a per-layer transcoder (PLT) architecture following a similar procedure to~\cite{ameisen2025circuit, dunefsky2024transcoders}. In this configuration, each MoE block is assigned a standalone transcoder trained in isolation. 

The PLT uses an encoder identical to the CLT (Equation \ref{eq:encoder}) to map the residual stream $\mathbf{x}^{\ell}$ into sparse activations $\mathbf{a}^{\ell}$. However, unlike the CLT, the PLT reconstruction $\hat{\mathbf{y}}^{\ell}$ is strictly local, utilizing only the latents present at layer $\ell$:
\begin{equation}
    \hat{\mathbf{y}}^{\ell} = \mathbf{W}_{\text{dec}}^{\ell \rightarrow \ell} \mathbf{a}^{\ell} + \mathbf{b}_{\text{pre}}^{\ell},
    \label{eq:decoder_plt}
\end{equation}
where we use the notation $\mathbf{W}_{\text{dec}}^{\ell \rightarrow \ell}$ for consistency. The PLT contains only one decoder matrix per layer, thereby lacking the capacity to utilize information from preceding layers. We then train the PLT via the optimization objective $\mathcal{L}_{\text{PLT}} = \mathcal{L}_{\text{MSE}} + \alpha \mathcal{L}_{\text{aux}}$, which mirrors the CLT training process but is applied to $L$ independent models.

\section{Replacement Models}
\label{app:replacement}

We utilize replacement models that approximate the ProGen3 forward pass using our CLT and PLT. These models are adapted from the ProtoMech framework~\cite{tsui2026protomech}. We provide a brief summary of each replacement model here.

\subsection{Transformer Notation}

We first set mathematical notation for transformers, following a similar notation to~\cite{elhage2021mathematical, dunefsky2024transcoders}, noting that ProGen3 is uses a sparse MoE architecture. This notation is meant to illustrate the replacement models, rather than provide a comprehensive overview of the ProGen3 architecture. 

Let $\mathbf{x}^{\ell}_{\text{pre}}$ represent the residual stream at the start of layer $\ell$. The attention sublayer produces an update added to the stream to yield $\mathbf{x}^{\ell}$:
\begin{equation}
    \mathbf{x}^{\ell} = \mathbf{x}^{\ell}_{\text{pre}} + \sum_{h \in H_{\ell}} h(\mathbf{x}^{\ell}_{\text{pre}}),
\end{equation}
where $H_{\ell}$ is the set of attention heads. The subsequent sublayer, denoted here as $\text{MoE}^{\ell}(\cdot)$, generates the output $\mathbf{y}^{\ell}$:
\begin{equation}
    \mathbf{y}^{\ell} = \text{MoE}^{\ell}(\mathbf{x}^{\ell}).
\end{equation}
The forward pass concludes with the update $\mathbf{x}^{\ell+1}_{\text{pre}} = \mathbf{x}^{\ell} + \mathbf{y}^{\ell}$.

\subsection{Overview of Replacement Models}
\label{app:replacement_models}

As reconstruction errors tend to compound during sequential inference~\cite{ameisen2025circuit}, we employ three distinct replacement strategies to analyze the trade-off between fidelity and computational realism.

\textbf{Direct Replacement.} This model calculates the final layer reconstruction $\hat{\mathbf{y}}^L$ using the ground-truth residual stream activations $\mathbf{x}^{\ell}$ as input at every layer. Because it bypasses the propagation of error through the network, it provides an upper bound on the CLT’s representational fidelity. Notably, this model is only viable for the CLT architecture. In a PLT, where decoders are strictly local, the ablation of a latent at $\ell < L$ would have no measurable effect on the final output if the ground-truth residual stream is provided at layer $L$.

\textbf{Sequential Replacement.} This variant aims to replicate the forward pass while minimizing error accumulation by utilizing ground-truth attention outputs. At each layer, we approximate $\mathbf{y}^{\ell}$ as $\hat{\mathbf{y}}^{\ell}$ and compute the subsequent residual stream as $\hat{\mathbf{x}}^{\ell+1}_{\text{pre}} = \mathbf{x}^{\ell} + \hat{\mathbf{y}}^{\ell}$. This model allows us to fairly compare the relative sparse recovery power of CLT and PLT architectures while keeping the attention mechanism constant. This replacement model is tested in the main text.

\textbf{Full Replacement.} Following \cite{merullo2025replicating}, this model replaces $\mathbf{y}^{\ell}$ with its transcoder reconstruction at every layer and relies solely on the initial input $\mathbf{x}^{0}_{\text{pre}}$. Unlike the sequential model, the full replacement model utilizes approximated attention outputs $h(\hat{\mathbf{x}}^{\ell}_{\text{pre}})$, allowing error to propagate through both the MoE and attention sublayers. This represents the most challenging setting for circuit discovery but offers the most faithful approximation of an entirely autonomous replacement model.

For the PLT baseline, we utilize only the sequential and full replacement strategies. Due to the lack of cross-layer connectivity, the direct replacement model has no meaningful analog in the PLT case.

\section{Experimental Setup}
\label{app:experimental-setup}

In this section, we detail our experimental setup for circuit discovery.

\subsection{Generation Circuit Discovery}
\label{app:generation_circuits}

\textbf{Data.} To construct tasks for generative circuit discovery, we utilize protein sequences from the Swiss-Prot database clustered at 30\% sequence identity obtained from~\cite{adams2025mechanistic}. From this dataset, we randomly sample 1000 sequences to serve as the basis for evaluation in both Causal Language Modeling (CLM) and Generalized Language Modeling (GLM) settings, ensuring that when generating with ProGen3, these sequences pass our low-quality filter check described in the main text. 

\textbf{Generative Tasks.} We evaluate two distinct generative modalities. For the CLM task, we provide the model with a prefix consisting of the first 80\% of a ground-truth Swiss-Prot sequence and task the model with autoregressively generating the remaining 20\%. For the GLM task, we perform span infilling by masking a segment of the sequence. The length of the masked span is determined by sampling from two Gaussian distributions, $\mathcal{N}(10, 5)$ and $\mathcal{N}(30, 10)$, with equal probability. During generation, we employ top-$p$ sampling with $p=0.95$ and a temperature $T=0.5$ to accentuate the distributional characteristics of the model. To isolate the reconstruction capability of the circuits and prevent temporal error accumulation, we provide the replacement model with ground-truth ProGen3 activations from the preceding step at each generative iteration.

\textbf{Circuit Discovery.} The objective of the generation circuit is to identify the minimal subset of latents required to faithfully recover ProGen3’s probability distribution. First, we compute the mean KL divergence achieved by the full CLT or PLT model (using all latents) relative to the original ProGen3 logits, denoted as $KL_{\text{all}}$. We then define the target threshold $\theta$ as $1.2 \times KL_{\text{all}}$, representing the point at which the circuit provides an approximation of the full model's generative behavior.

To identify these circuits, we utilize an iterative greedy search based on gradient attribution. For each sequence, we compute an attribution score $\Delta_i^{\ell}$ for every latent $i$ at every layer $\ell$ with respect to the mean KL divergence between the circuit and ProGen3. In these generative tasks, the KL divergence is calculated exclusively over the logits of the generated tokens:
\begin{equation}  
\Delta_i^\ell = \left| a_i^\ell \cdot \frac{\partial KL_{\text{gen}}}{\partial a_i^\ell} \right|.
\label{eq:attribution_gen}
\end{equation}
This provides a ranking of the latents that influenced the generation of the sequence. We then construct the circuit by ranking latents in descending order of attribution and incrementally adding them in batches of 32. At each step, we evaluate the mean KL divergence of the resulting sparse reconstruction on the generated sequence, continuing the process until the target threshold $\theta$ is reached, or the circuit size reaches a maximum of 1000 latents.

\subsection{Zero-shot Function Prediction Circuit Discovery}
\label{app:function_circuits}

\begin{table*}[htbp]
\vspace{-0.2cm}
\caption{Summary of DMS assays used.}
\label{function-circuit-table}
\vspace{0.2cm}
\label{tab:dms}
\centering
\resizebox{\textwidth}{!}{%
\begin{tabular}{lcccccr}
\toprule
\textbf{DMS} & \textbf{Description} & \textbf{Function Tested} \\
\midrule
A4\_HUMAN\_Seuma~\cite{seuma2022atlas}  & Amyloid-beta peptide           & Aggregation \\
CAPSD\_AAV2S\_Sinai~\cite{sinai2021generative} & Adeno-associated virus capsid             & Viral production  \\
F7YBW8\_MESOW\_Ding~\cite{ding2024protein}  & Antitoxin ParD3           & Growth enrichment  \\
GFP\_AEQVI\_Sarkisyan~\cite{sarkisyan2016local}   & Green fluorescent protein & Fluorescence  \\
GRB2\_HUMAN\_Faure~\cite{faure2022mapping} & C-terminal SH3 domain of GRB2  & Yeast growth \\
RASK\_HUMAN\_Weng\_abundance~\cite{weng2023energetic} & KRAS             & Yeast growth \\
SPG1\_STRSG\_Olson~\cite{olson2014comprehensive}  & IgG-binding domain of protein G  & Binding \\
YAP1\_HUMAN\_Araya~\cite{araya2012fundamenta;} & hYAP65 WW domain             & Peptide binding \\
\bottomrule
\end{tabular}%
}
\end{table*}

\textbf{Data.} To identify function circuits, we use 8 DMS assays from ProteinGym~\cite{notin2023proteingym}. These protein selections ensure robust evaluation across a variety of functions (Table~\ref{tab:dms}). All of these DMS assays contain both single and multiple mutants, and we sample randomly from all mutation types.

\textbf{Training Splits}. For each DMS assay, we randomly sample 256 mutants, where half the sequences come from randomly sampling sequences with \texttt{DMS\_score\_bin = 1} and the other half from sampling sequences with \texttt{DMS\_score\_bin = 0} to ensure appropriate coverage of the data for zero-shot function prediction. Due to computational constraints, we then sample 1000 random mutants from the remaining sequences to use as our test set with the same split of functional sequences. We repeat this procedure five times for five folds, and each CLT and PLT model receives the same training and test set for a given fold.

\textbf{Circuit Discovery.} First, we define $m_{\text{clean}}$ to be the Spearman correlation obtained from ProGen3-112M using the 256 training mutants. Following a similar procedure to~\cite{nainani2025mechanistic, tsui2026protomech}, we define a function circuit as the smallest subset of latents required to achieve a comparable zero-shot Spearman correlation to $m_{\text{clean}}$, where a comparable score is 70\% of $m_{\text{clean}}$. However, there are cases in which adding all latents does not achieve 70\% of the score due to lossy reconstruction, so we compute $m_{\text{all}}$ for a given model as the Spearman correlation with all the latents active. We then adjust the target to the maximum possible performance achievable: $\theta = \min(0.7 \times m_{\text{clean}}, m_{\text{all}})$.

To identify the subset of latents for a circuit, we utilize an iterative greedy search similar to Appendix~\ref{app:generation_circuits}. For each sequence, we compute an attribution score $\Delta_i^{\ell}$ for every latent $i$ at every layer $\ell$, with respect to the mean KL divergence between the final logit matrices of the replacement model and ProGen3, denoted as $KL_{model}$. For a given sequence, we compute the attribution score as a product of the latent's activation $a_i^{\ell}$ and its gradient with respect to $KL_{model}$:
\begin{equation} 
\Delta_i^\ell = \left| a_i^\ell \cdot \frac{\partial KL_{model}}{\partial a_i^\ell} \right|.
\label{eq:attribution}
\end{equation}
$\Delta_i^{\ell}$ estimates the contribution of each latent to the mean KL divergence. We sum these attribution scores across all 256 sequences in the training set to obtain a global attribution ranking for each latent.

To iteratively construct the circuit, we rank the latents by descending attribution score and add the top-ranked latents in steps of 32 at a time, up to a maximum of 1000 latents. At each step, we compute the zero-shot Spearman correlation $m_\text{circuit}$ on the training sequences using the sparse reconstruction derived from the top latents, and continue until $m_\text{circuit} \geq \theta$ or until 1000 latents.

\section{Visualization tool}
\label{app:viz_tool}

We provide the visualization tool alongside our paper submission. The platform enables users to interactively explore the circuit structures presented in the paper and analyze new protein sequences. Users can either select from pre-loaded examples or upload custom circuit data by providing the following files:
\vspace{-4mm}
\begin{enumerate}
    \item \texttt{activation\_indices.json}: Contains latents (specified by layer and latent index) that activate at each position in the sequence.
    \vspace{-2mm}
    \item \texttt{generation.fasta}: Contains the protein sequence in FASTA format, including a \texttt{<CLM>} or \texttt{<GLM>} marker that separates the prompt from the generated portion.
    \vspace{-2mm}
    \item \texttt{top\_activations.json}: Contains the top-activating sequences for each latent.
    \vspace{-2mm}
    \item \texttt{logits.npy} (CLM/GLM only): Contains the model's per-position logits over the vocabulary at each generated position.
    \vspace{-2mm}
    \item \texttt{virtual\_weights.json} (optional): Contains all virtual weights between pairs of latents.
\end{enumerate}

\paragraph{Analysis Modes.} The tool supports three analysis modes, automatically detected from \texttt{generation.fasta}:
\begin{itemize}
    \item \textbf{CLM (Causal Language Modeling).} The prompt contains a \texttt{<CLM>} marker indicating where left-to-right autoregressive generation begins, and \texttt{>output} contains the generated residues. The visualization shows the marker boundary in the sequence axis and, for each generated position, the model's top-ranked next-token predictions sourced from \texttt{logits.npy}.
    \vspace{-1mm}
    \item \textbf{GLM (Generative/Span Language Modeling).} Same file layout as CLM, but the marker is \texttt{<GLM>} and generation corresponds to span infilling rather than left-to-right continuation. Downstream visualization is identical to CLM (top-token logits per generated position), differing only in what the generated span semantically represents.
    \item \textbf{Zero-shot.} The prompt is a complete protein sequence with no marker, and the \texttt{>output} record stores a single scalar (e.g., the model's log-likelihood or a fitness score). The visualization displays latent activations across the entire input sequence and reports the score; no per-position logits are shown because nothing is generated.
\end{itemize}

We describe the expected structure of each file below.

\paragraph{\texttt{generation.fasta}.} Two-record FASTA. The \texttt{>prompt} record contains the input sequence, optionally including a \texttt{<CLM>} or \texttt{<GLM>} marker that indicates where the generated portion begins. The \texttt{>output} record contains either the generated residues (CLM/GLM mode) or a single zero-shot score (zero-shot mode).
\begin{verbatim}
>prompt
MVKQVDFAEVKLSEKFLGAGSGGAVRKATFQNQEIAVKIFDFLEETIKKNAE...<CLM>
>generated_output
D
\end{verbatim}

\paragraph{\texttt{activation\_indices.json}.} Flat JSON array whose entries are \texttt{[layer, position, value, latent\_index]} tuples, listing every latent that activates at each position of the sequence.
\begin{verbatim}
[
  [0, 7,  6.787, 2692],
  [0, 34, 7.830, 1521],
  [7, 2,  3.101, 2747],
  ...
]
\end{verbatim}

\paragraph{\texttt{top\_activations.json}.} Nested object keyed by layer and latent index. Each latent maps to a list of its top-activating reference proteins, including a per-residue activation profile and metadata.
\begin{verbatim}
{
  "num_layers": 10,
  "family": "GRB2",
  "layers": {
    "0": {
      "1521": [
        {
          "Score": 9.31,
          "Activations": [0.85, 0.0, ..., 9.31, ...],
          "Entry Name": "RS4_PICTO",
          "Entry": "Q6KZP7",
          "Protein names": "Small ribosomal subunit protein uS4",
          "Sequence": "MGDQKFQRKKYSTPRHPWEKDR...",
          "seq_len": 183
        }, ...
      ]
    }, ...
  }
}
\end{verbatim}

\paragraph{\texttt{logits.npy} (CLM/GLM only).} A 2D \texttt{float32} NumPy array of shape \texttt{[num\_generated, vocab\_size]}. Row \(i\) contains the model's logits over the vocabulary at the \(i\)-th generated position.
\begin{verbatim}
# numpy array
shape:  (num_generated, vocab_size) 
dtype:  float32
row[0]: [-16.75, -16.75, -14.56, ..., -16.75]
\end{verbatim}

\paragraph{\texttt{virtual\_weights.json} (optional).} Flat JSON array of edge tuples \texttt{[src\_pos, src\_layer, src\_feature, tgt\_pos, tgt\_layer, tgt\_feature, weight]} encoding the virtual weight between a source latent at one position and a target latent at another.
\begin{verbatim}
[
  [1, 7, 2227, 1, 9, 957, -0.2195],
  ...
]
\end{verbatim}

\begin{figure*}[t!]
\centering
\includegraphics[width=0.9\textwidth]{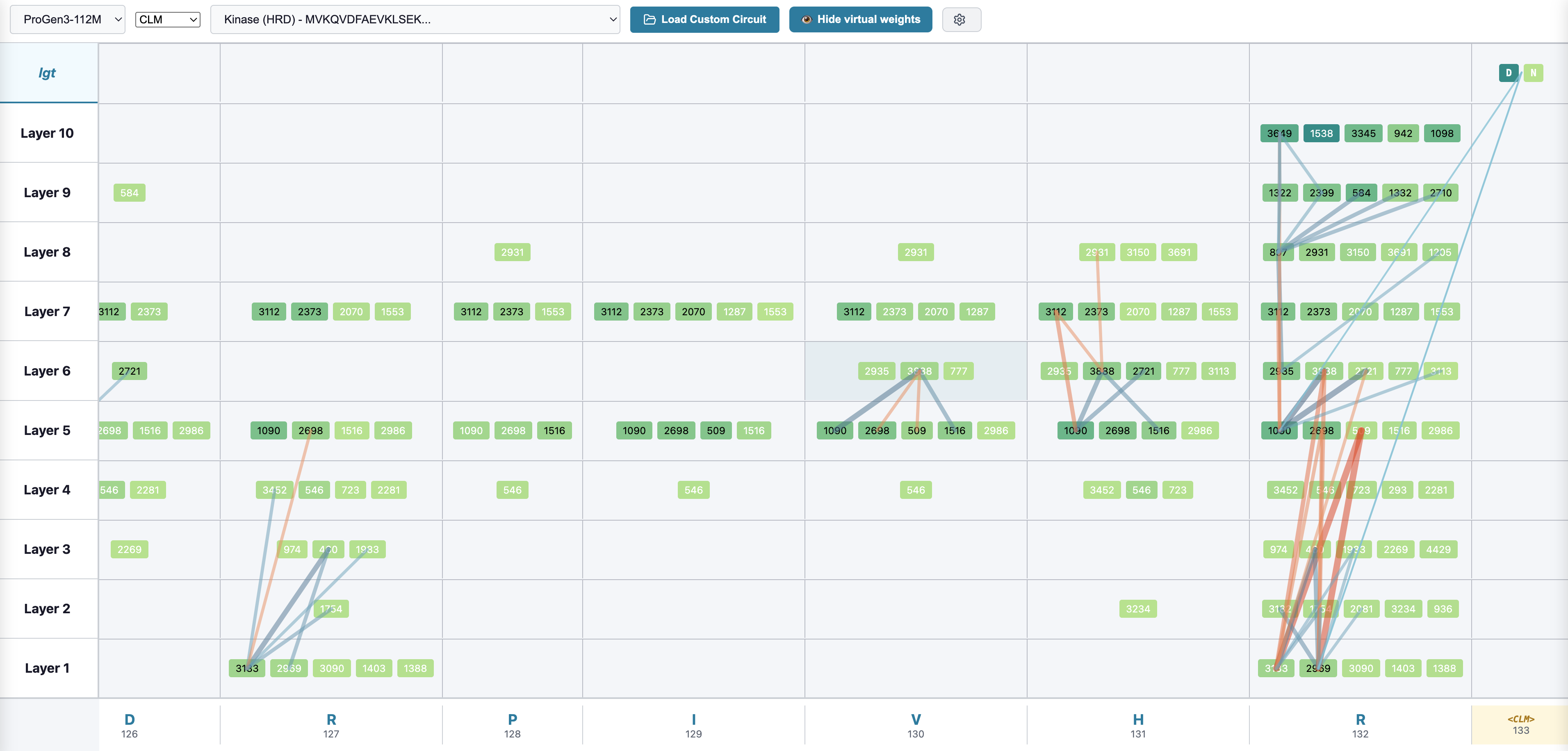}
\caption{\textbf{Main interface.} The grid view displays latent activations organized by CLT layer (rows) and sequence position (columns). Each colored box represents an active latent and is labeled with its index; the shade of green encodes activation magnitude. When virtual weights are enabled, edges connect latents across layers: blue edges indicate positive weights (excitation) and orange edges indicate negative weights (inhibition). The bottom axis shows the protein sequence by single-letter code and position, with the \texttt{<CLM>}/\texttt{<GLM>} marker rendered at the prompt/generation boundary in CLM/GLM mode. The top \texttt{lgt} row (CLM/GLM only) lists the model's top predicted tokens at each generated position. The toolbar (top) provides controls for selecting the model, analysis mode, and example circuit, loading custom data, toggling virtual-weight edges, and accessing settings.}
\label{fig:website_grid_view}
\end{figure*}

\paragraph{Main Interface.} The primary view displays a grid layout where rows correspond to CLT layers and columns correspond to sequence positions (Figure~\ref{fig:website_grid_view}). Each cell contains the latents that activate at that position, rendered as colored boxes with the latent index inside; activation strength is encoded by color intensity. When virtual weights are enabled, edges connect latents across layers, with blue indicating positive weights and orange indicating negative weights. The protein sequence is displayed along the bottom axis, with each residue labeled by its single-letter amino-acid code and absolute position; in CLM/GLM mode, the marker (\texttt{<CLM>} or \texttt{<GLM>}) is rendered at the boundary between the prompt and the generated portion. For CLM/GLM analyses, an additional \texttt{lgt} row at the top of the grid shows the model's top predicted tokens (sourced from \texttt{logits.npy}) at each generated position. The toolbar at the top of the page exposes the model selector, analysis-mode dropdown, example circuit picker, a \emph{Load Custom Circuit} button for uploading user-provided files, a toggle to show/hide virtual-weight edges, and a settings menu.

\begin{figure*}[t!]
\centering
\includegraphics[width=0.9\textwidth]{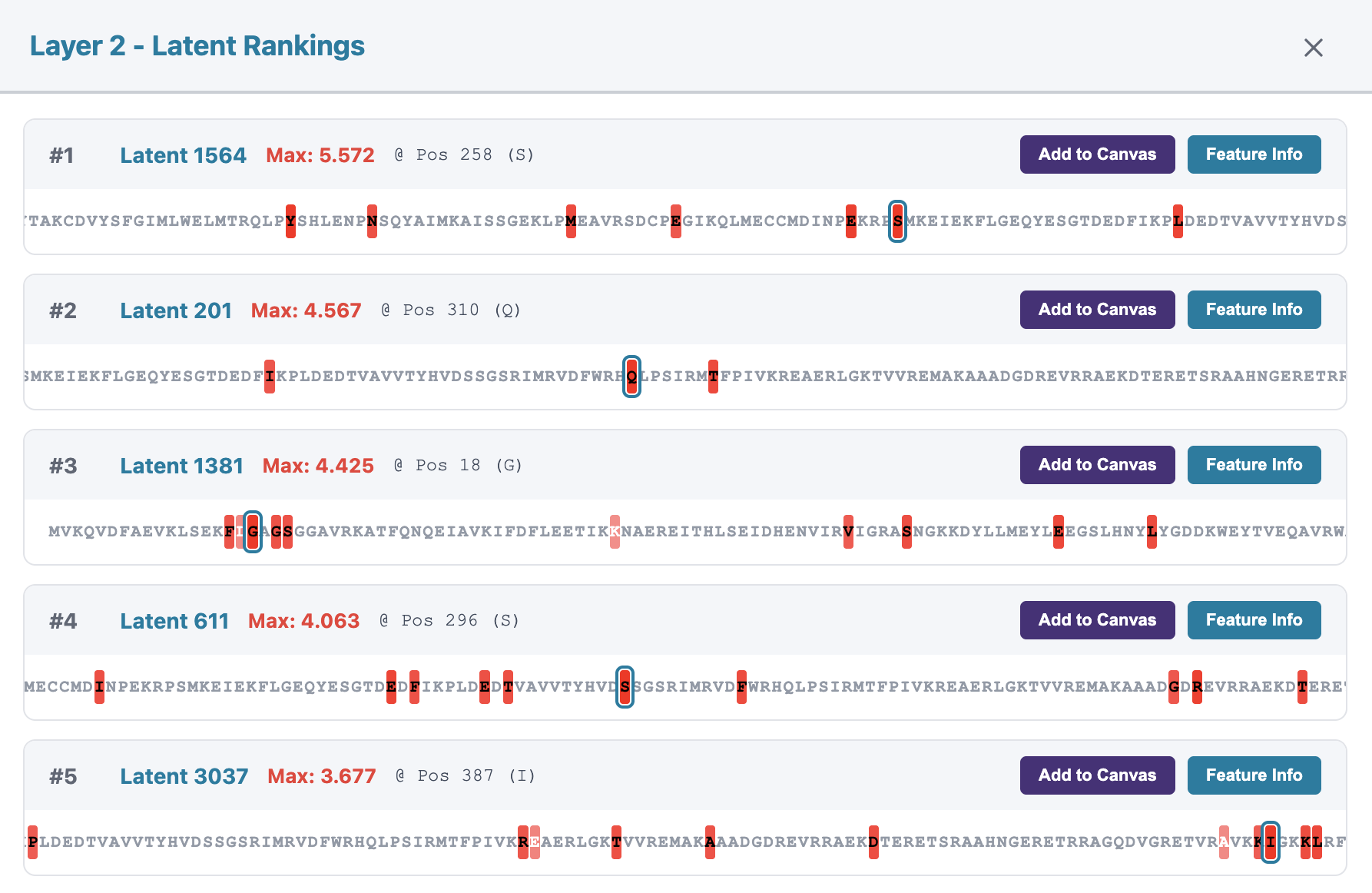}
\caption{\textbf{Layer-wise latent rankings.} Clicking on a layer label opens this panel, showing the top-activating latents for that layer ranked by maximum activation value. Each entry displays the latent index, maximum activation score, the sequence position of peak activation, and a sequence window centered on the activation site (highlighted in red). The ``Add to Canvas'' button enables quick circuit construction, while ``Feature Info'' opens the detailed latent information panel.}
\label{fig:latent_ranking_tab}
\end{figure*}

\paragraph{Exploring Latent Rankings.} Clicking on a layer label (e.g., ``Layer 2'') opens the Latent Rankings panel, which displays the top-activating latents for that layer ranked by their maximum activation value (Figure~\ref{fig:latent_ranking_tab}). Each entry shows the latent index, maximum activation score, the sequence position where maximum activation occurs, and a sequence context window with the activation site highlighted. Users can directly add latents to the canvas or access detailed feature information from this panel.

\begin{figure*}[t!]
\centering
\includegraphics[width=0.9\textwidth]{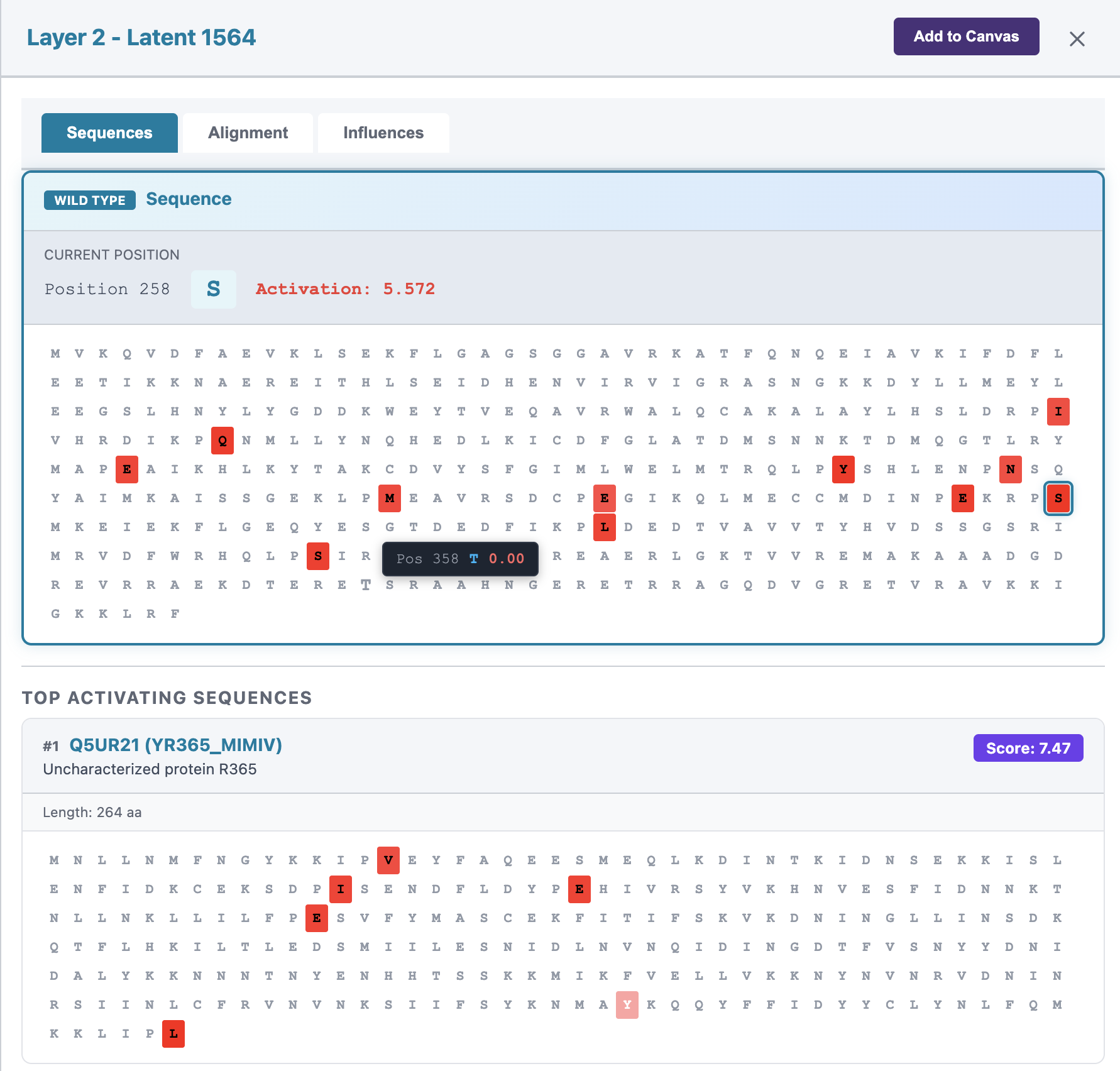}
\caption{\textbf{Latent Information Panel: Sequences tab.} The Sequences tab provides detailed activation information for a selected latent. The top section shows the input sequence with activation intensities displayed as a heatmap, where darker colors indicate stronger activation. The current position, amino acid, and activation value are displayed above the sequence. The bottom section lists top-activating sequences from Swiss-Prot, each with its UniProt identifier, protein name, and activation score, enabling users to identify sequence patterns that maximally activate the latent.}
\label{fig:latent_details_tab}
\end{figure*}

\begin{figure*}[t!]
\centering
\includegraphics[width=0.9\textwidth]{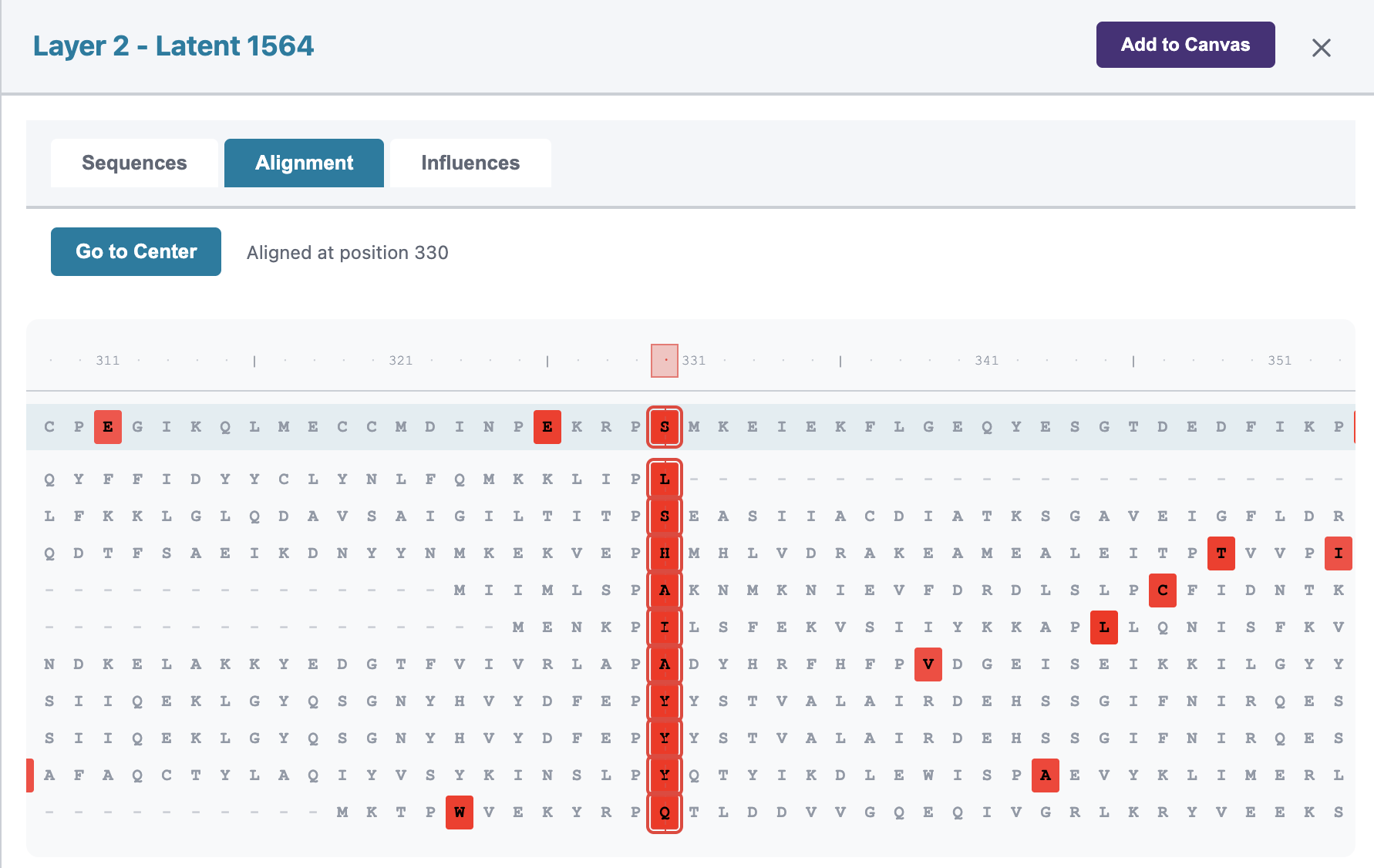}
\caption{\textbf{Latent Information Panel: Alignment tab.} The Alignment tab displays an alignment between the input sequence (top row) and top-activating sequences from Swiss-Prot. Activation intensities are shown as colored overlays, with red indicating high activation. The ``Go to Center'' button navigates to the position of maximum activation of wild type and top activating sequences. This view facilitates identification of conserved sequence motifs and structural features recognized by the latent.}
\label{fig:latent_alignment_tab}
\end{figure*}

\begin{figure*}[t!]
\centering
\includegraphics[width=0.9\textwidth]{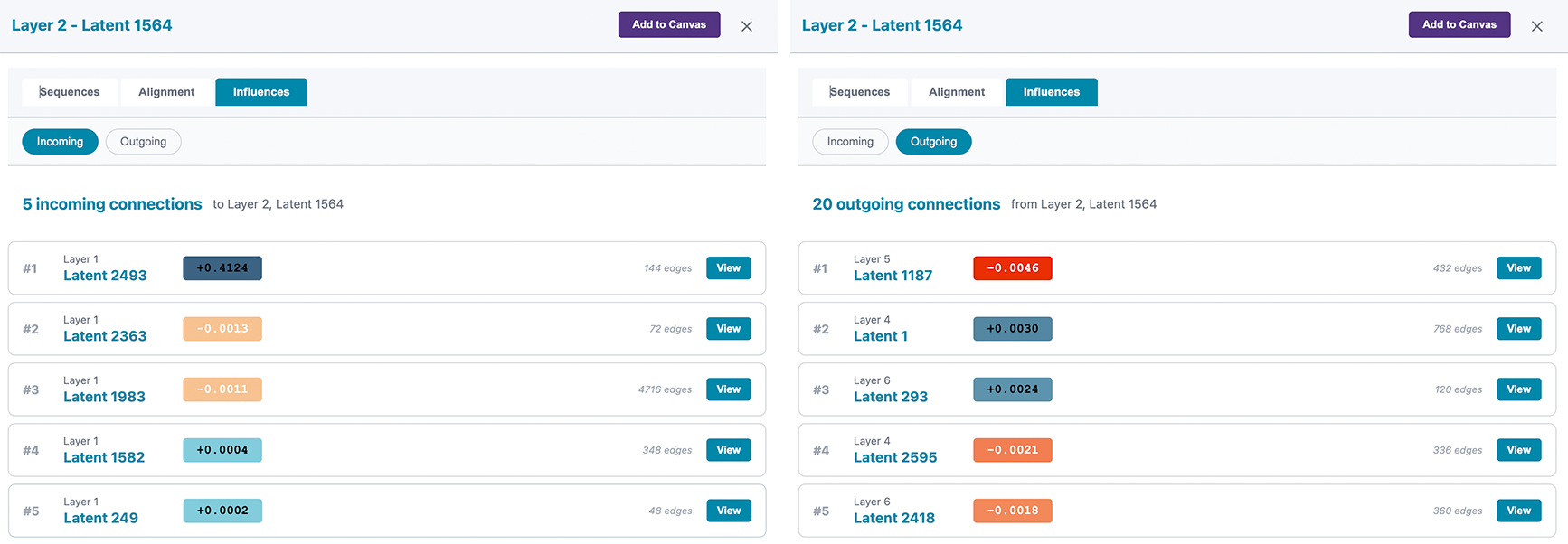}
\caption{\textbf{Latent Information Panel: Influences Tab.} The Influences tab reveals the circuit connectivity of a selected latent. The left panel shows incoming connections from earlier layers, while the right panel shows outgoing connections to later layers. Each connection displays the source/target layer and latent index, the virtual weight (green for positive, orange for negative), and the number of sequence positions where this edge is active.}
\label{fig:latent_influences_tab}
\end{figure*}

\paragraph{Latent Information Panel.} Clicking on any latent opens a detailed information panel with three tabs (Figures~\ref{fig:latent_details_tab}--\ref{fig:latent_influences_tab}):

\begin{itemize}[leftmargin=10pt]
    \vspace{-4mm}
    \item \textit{Sequences} (Figure~\ref{fig:latent_details_tab}): Displays the input sequence with a heatmap overlay showing activation intensities across all positions. The current position's amino acid and activation value are shown at the top. Below, the top-activating sequences from Swiss-Prot are listed with their activation scores, allowing users to identify common patterns that drive latent activation.

    \vspace{-2mm}
    \item \textit{Alignment} (Figure~\ref{fig:latent_alignment_tab}): Presents an alignment view where the input sequence is aligned with the top-activating sequences from Swiss-Prot. Activation intensities are overlaid as colored highlights, enabling users to identify conserved motifs and structural features that the latent has learned to recognize.

    \vspace{-2mm}
    \item \textit{Influences} (Figure~\ref{fig:latent_influences_tab}): Lists all incoming connections (from earlier layers) and outgoing connections (to later layers) for the selected latent. Each connection displays the source/target layer and latent index, the virtual weight magnitude (positive in green, negative in orange), and the number of sequence positions where this connection is active. This view enables systematic exploration of how information flows through the circuit.
\end{itemize}

\begin{figure*}[t!]
\centering
\includegraphics[width=0.9\textwidth]{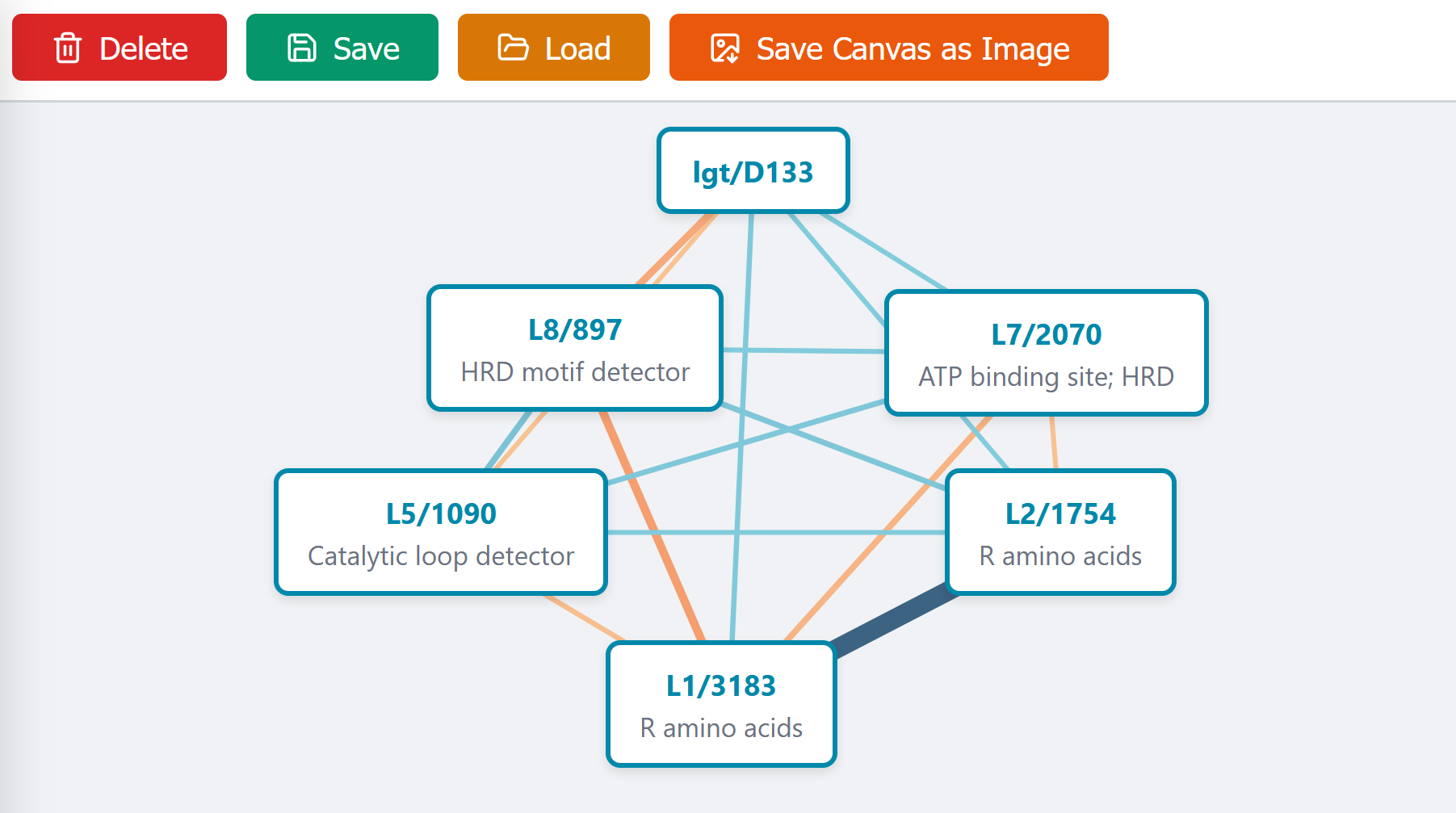}
\caption{\textbf{Interactive Circuit Canvas.} The canvas provides a workspace for constructing and annotating circuit diagrams. Nodes represent latents, labeled as Layer/Index (e.g., ``L8/897'', ``L7/2070''), and can optionally be annotated with a descriptive name documenting their hypothesized function (e.g., ``HRD motif detector'', ``ATP binding site'', ``Catalytic loop detector''); annotations are added by right-clicking a node. In CLM/GLM mode, the canvas additionally supports special \texttt{lgt/<token><pos>} nodes (e.g., ``lgt/D133'') that represent a model-predicted top token at a generated position, allowing virtual-weight relationships between upstream latents and the generation logits to be inspected directly. Edges represent virtual weights between latents, with blue indicating positive weights and orange indicating negative weights; edge thickness is proportional to weight magnitude. The toolbar provides functions for deleting selected nodes, saving and loading canvas layouts, exporting the canvas as an SVG image, and toggling fullscreen mode.}
\label{fig:supp_canvas}
\end{figure*}

\paragraph{Building and Analyzing Circuits on the Canvas.} The canvas provides an interactive workspace for constructing and visualizing circuit diagrams (Figure~\ref{fig:supp_canvas}). Users can add latents to the canvas by right-clicking on any latent in the grid view or using the ``Add to Canvas'' button in the Latent Info panel. Once on the canvas, nodes can be freely repositioned and annotated with descriptive labels (e.g., ``Glycine-rich loop'', ``ATP-binding site'') to document hypothesized functions. Virtual weight edges are automatically drawn between canvas nodes, with edge color indicating sign (blue for positive, orange for negative) and thickness proportional to magnitude. The ``Filter Virtual Weights'' slider controls the minimum weight threshold for displayed edges. Canvas layouts can be saved and loaded for reproducibility, and exported as images for publication.

\begin{figure*}[t!]
\centering
\includegraphics[width=0.9\textwidth]{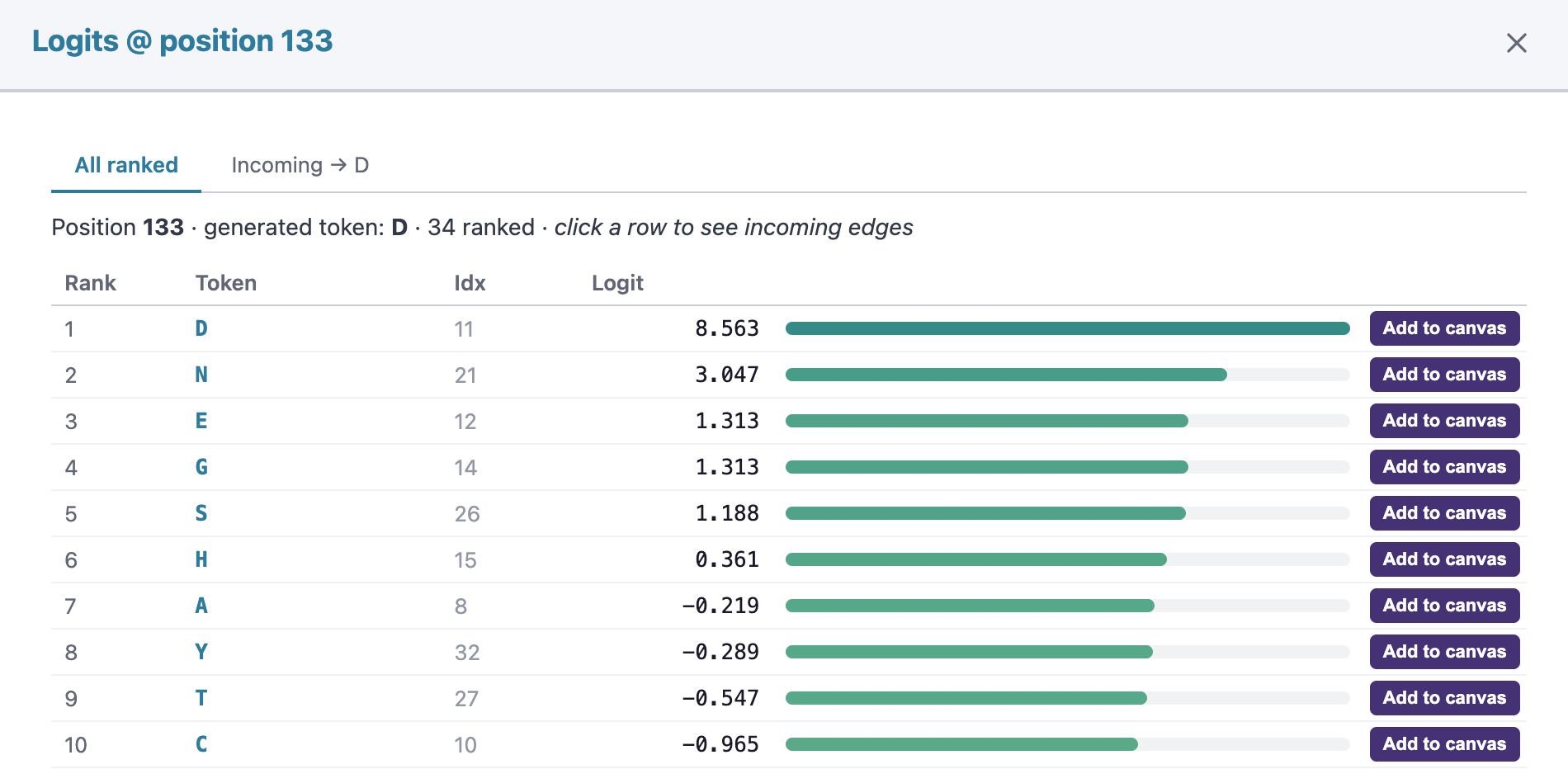}
\caption{\textbf{Logit Panel.} Detailed view of the model's predictions at a single generated position (CLM/GLM only), opened by clicking the corresponding column header in the \texttt{lgt} row of the main grid. The header reports the position index and the actually generated token (here, position 133 with token \textbf{D}). The \emph{All ranked} tab lists every in-vocab token ranked by logit value, showing its rank, single-letter symbol, vocabulary index, and numeric logit, with an inline horizontal bar visualizing magnitude relative to the top entry. The \emph{Incoming $\rightarrow$ D} tab restricts the view to virtual-weight contributions flowing into the generated token, helping identify which upstream latents drove the prediction. The \emph{Add to canvas} button on each row inserts the corresponding \texttt{lgt/<token><pos>} node into the interactive canvas, allowing the user to inspect virtual-weight pathways between upstream latents and the predicted token.}
\label{fig:logit_panel}
\end{figure*}

\clearpage

\section{Additional Experimental Results}
\label{app:expt_results}

In this section, we detail our experimental results that did not fit the main text.

\begin{figure*}[t!]
\vspace{-0cm}
\centering
\includegraphics[width=1\textwidth]{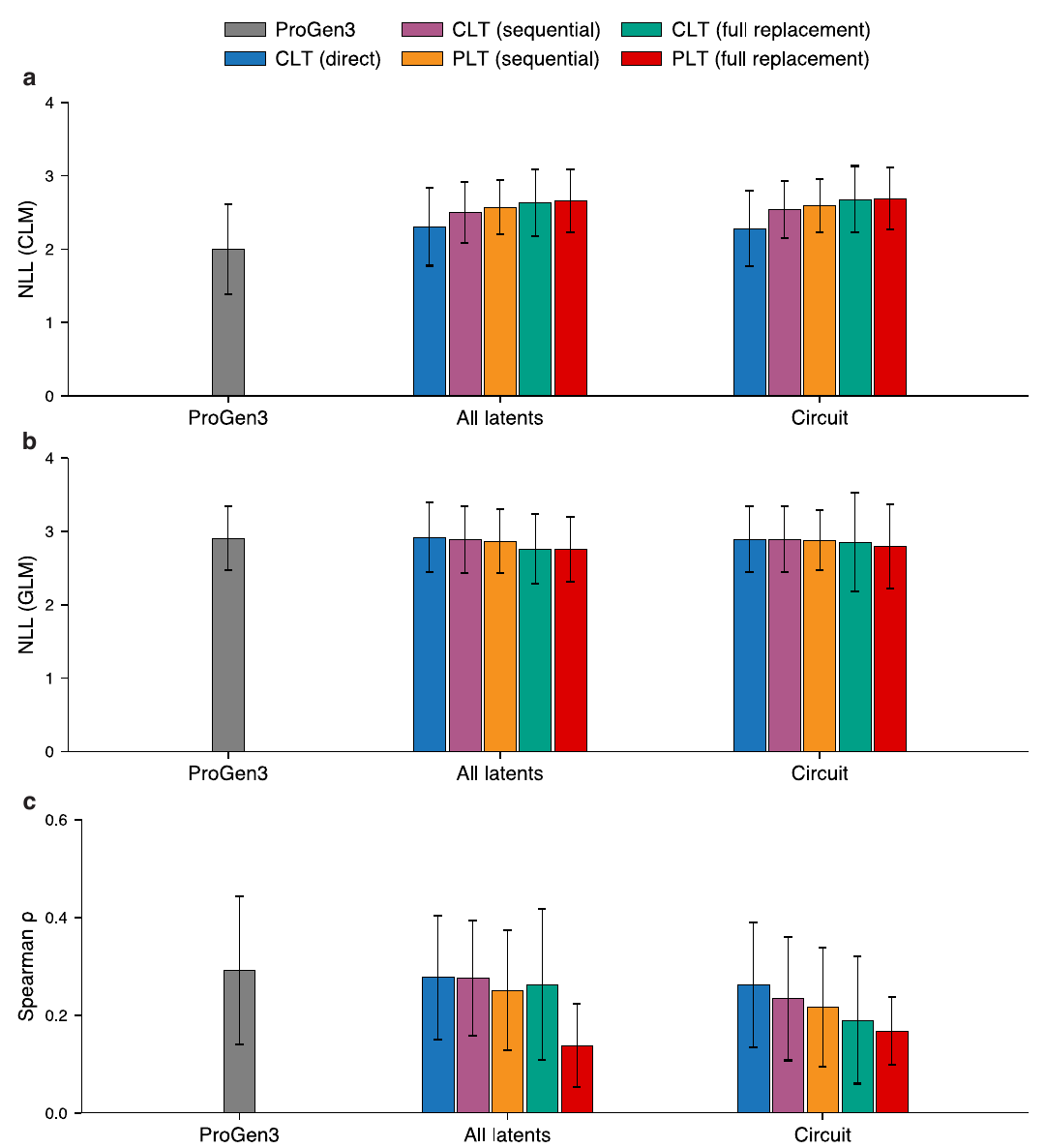}
\vspace{-0.3cm}
\caption{Performance of all replacement models on generation (CLM and GLM) and zero shot fitness prediction.
} 
\label{fig:supp_circuit_performance}
\end{figure*}

\subsection{Circuit Discovery}

Fig.~\ref{fig:supp_circuit_performance} illustrates the performance of all replacement models across \textbf{a,} CLM generation, \textbf{b,} GLM generation, and \textbf{c,} zero-shot fitness prediction. In the CLM and zero-shot fitness tasks, we observe a consistent hierarchy in performance. The CLT direct replacement model achieve the highest fidelity to the base ProGen3 model, as it bypasses the accumulation of reconstruction errors inherent in layer-wise sequential operations. When controlling for the specific replacement strategy, the CLT architecture consistently outperforms the PLT baseline. We also observe a significant performance gap between the sequential and full replacement models. This is consistent with observations in~\cite{ameisen2025circuit}, as error propagation becomes significantly more magnified with reconstructing the attention mechanism at every layer. In the GLM task, we observe that all the replacement models provide decent approximations to the original model. As discussed in Section~\ref{sec:generation_results}, we attribute this to 112M model's lack of capability in generating functionally plausible sequences in the infilling objective.

\begin{figure*}[t!]
\centering
\includegraphics[width=1\textwidth]{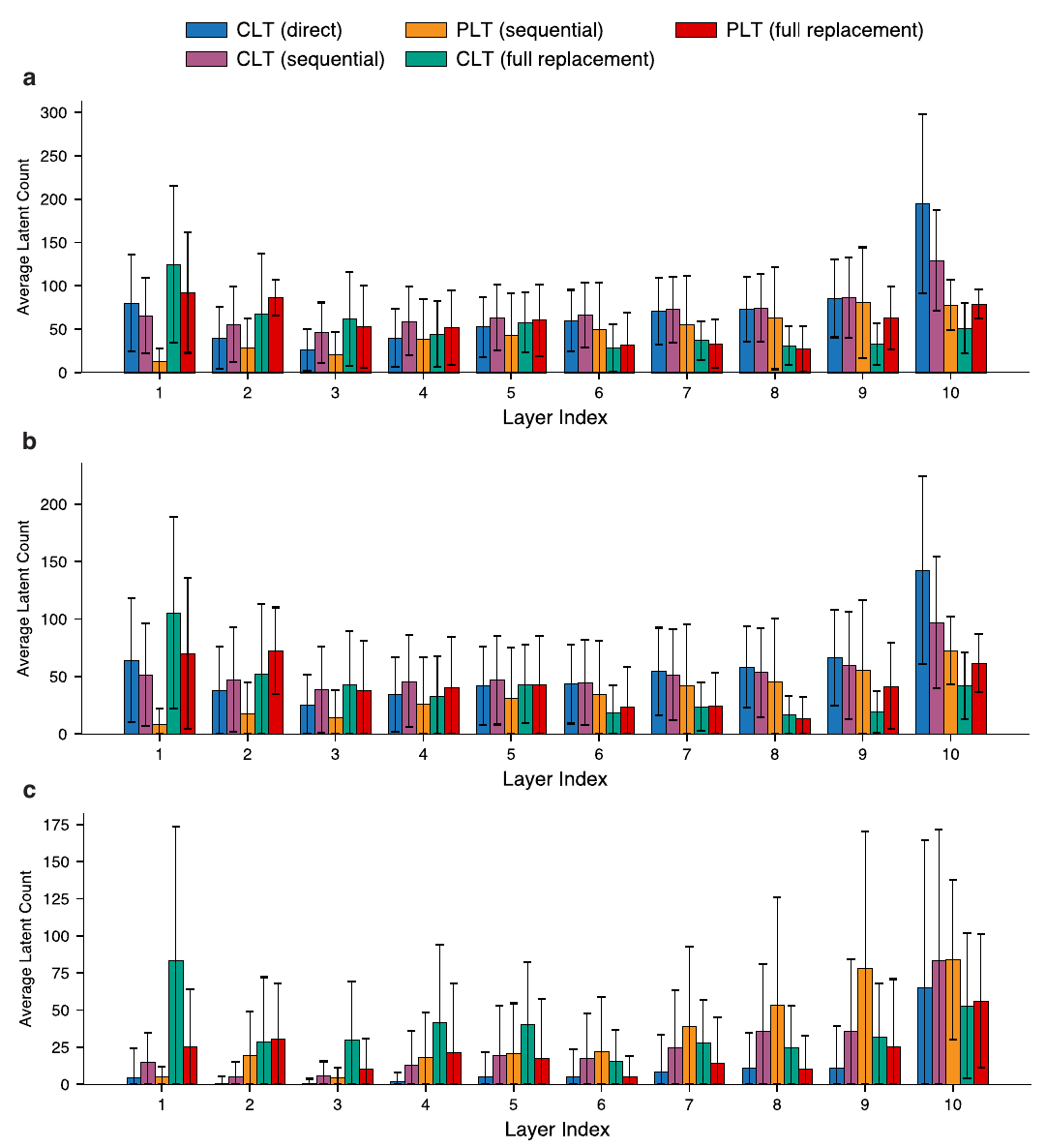}
\caption{Average number of latents in circuits across all replacement models for {\bf a}, CLM generation, {\bf b}, GLM generation, and {\bf c}, zero-shot fitness prediction.}
\label{fig:supp_latents}
\end{figure*}

Fig.~\ref{fig:supp_latents} details the average number of latents comprising the circuits for \textbf{a,} CLM generation, \textbf{b,} GLM generation, and \textbf{c,} zero-shot fitness prediction. Across all tasks, we observe a distinct U-shaped distribution of latent density: latent usage is most concentrated in the final layers, followed by the initial layers, while the middle layers exhibit the highest degree of sparsity. This distribution is consistent with findings in~\cite{tsui2026protomech}. We attribute this trend to the levels of information in the model. The initial layers appear to capture a vast array of fundamental biochemical patterns and local amino acid dependencies, necessitating a high volume of specific features. Similarly, the final layers appear to host specialized motif detectors and task-specific features critical for precise generation and scoring. In contrast, the middle layers appear to encode for higher-level structural motifs, which can be represented more efficiently with a significantly sparser subset of latents.

\subsection{Steering Results}

\begin{table*}[h]
\caption{Steering results using ProGen3, ProGenMech (sequential), and PLT (sequential) on four DMS assays. All variants were scored using the original ProGen3 average likelihood.}
\vspace{0.2cm}
\label{tab:ProGen_steering_results}
\centering
\resizebox{1 \textwidth}{!}{%
\begin{tabular}{lrccccc}
\toprule
\textbf{Method} & \textbf{DMS}
  & \textbf{Mean score $\uparrow$} & \textbf{Max score $\uparrow$}
  & \textbf{Top 10\% score $\uparrow$} & \textbf{Top 20\% score $\uparrow$} \\
\midrule
\textbf{ProGen3} & A4\_HUMAN\_Seuma\_2022 & -2.68 $\pm$ 0.02 & -2.63 & -2.64 $\pm$ 0.01 & -2.65 $\pm$ 0.01 \\
 & F7YBW8\_MESOW\_Ding\_2023 & -2.28 $\pm$ 0.01 & -2.25 & -2.26 $\pm$ 0.01 & -2.26 $\pm$ 0.01 \\
 & GRB2\_HUMAN\_Faure\_2021 & -2.47 $\pm$ 0.03 & -2.43 & -2.43 $\pm$ 0.00 & -2.43 $\pm$ 0.00 \\
 & YAP1\_HUMAN\_Araya\_2012 & -2.39 $\pm$ 0.02 & -2.35 & -2.36 $\pm$ 0.01 & -2.37 $\pm$ 0.01 \\
\cmidrule(lr){1-6}
\textbf{ProGenMech (sequential)} & A4\_HUMAN\_Seuma\_2022 & -2.73 $\pm$ 0.04 & -2.68 & -2.68 $\pm$ 0.00 & -2.68 $\pm$ 0.00 \\
 & F7YBW8\_MESOW\_Ding\_2023 & -2.32 $\pm$ 0.04 & -2.26 & -2.26 $\pm$ 0.00 & -2.27 $\pm$ 0.01 \\
 & GRB2\_HUMAN\_Faure\_2021 & -2.53 $\pm$ 0.11 & -2.43 & -2.43 $\pm$ 0.00 & -2.43 $\pm$ 0.00 \\
 & YAP1\_HUMAN\_Araya\_2012 & -2.53 $\pm$ 0.07 & -2.41 & -2.41 $\pm$ 0.00 & -2.42 $\pm$ 0.01 \\
\cmidrule(lr){1-6}
\textbf{PLT (sequential)} & A4\_HUMAN\_Seuma\_2022 & -2.67 $\pm$ 0.02 & -2.64 & -2.65 $\pm$ 0.00 & -2.65 $\pm$ 0.00 \\
 & F7YBW8\_MESOW\_Ding\_2023 & -2.33 $\pm$ 0.06 & -2.25 & -2.26 $\pm$ 0.00 & -2.27 $\pm$ 0.01 \\
 & GRB2\_HUMAN\_Faure\_2021 & -2.53 $\pm$ 0.08 & -2.42 & -2.42 $\pm$ 0.00 & -2.44 $\pm$ 0.02 \\
 & YAP1\_HUMAN\_Araya\_2012 & -2.42 $\pm$ 0.05 & -2.37 & -2.38 $\pm$ 0.00 & -2.38 $\pm$ 0.00 \\
\bottomrule
\end{tabular}%
}
\end{table*}

Following 
a similar steering methodology in~\cite{tsui2026protomech, ameisen2025circuit}, we evaluated the capacity of our replacement models to guide the generation of high-fitness protein variants. We selected the top four DMS assays where the original ProGen3 model exhibited the highest zero-shot Spearman correlation.

As shown in Table~\ref{tab:ProGen_steering_results}, we observe that the scores produced by the CLT and PLT sequential replacement models are distributionally similar to the original ProGen3 baseline across all metrics. A qualitative analysis of the generated output reveals a significant presence of low-complexity regions across all three methods (Fig.~\ref{fig:sequence_snippets}). While the replacement models successfully replicate the base model's performance, they also inherit its susceptibility to generate low-quality regions and fail to generate sequences with a substantially lower log-likelihood.

\begin{figure}[h]
\centering
\begin{tikzpicture}[
    font=\ttfamily\footnotesize,
    residue/.style={draw=black!15, minimum size=0.4cm, inner sep=0pt, outer sep=0pt, line width=0.2pt},
    label_node/.style={anchor=east, xshift=-0.2cm, font=\sffamily\scriptsize\bfseries}
]


\node[label_node] at (0, 0) {ProGen3};
\foreach \aa [count=\i] in {L,L,L,I,L,I,L,I,L,L,I,I,L,L,L,I,L,L,I,I} {
    \node[residue, fill=gray!10] at (\i*0.35, 0) {\aa};
}

\node[label_node] at (0, -0.45) {CLT (sequential)};
\foreach \aa [count=\i] in {A,F,V,I,I,I,L,I,I,L,L,L,L,G,I,G,L,L,L,I} {
    \node[residue, fill=blue!5] at (\i*0.35, -0.45) {\aa};
}

\node[label_node] at (0, -0.9) {PLT (sequential)};
\foreach \aa [count=\i] in {F,L,I,I,I,I,L,I,I,M,I,L,L,L,L,I,I,L,I,L} {
    \node[residue, fill=green!5] at (\i*0.35, -0.9) {\aa};
}

\end{tikzpicture}
\caption{Example of a generated portion from the \texttt{A4\_HUMAN\_Seuma\_2022} DMS assay. All three methods generate low-complexity regions.}
\label{fig:sequence_snippets}
\end{figure}


\end{document}